\RequirePackage{fix-cm}
\documentclass[twocolumn]{svjour3}          
\smartqed  

\usepackage[utf8]{inputenc} 
\usepackage[T1]{fontenc}    
\usepackage{url}            
\usepackage{booktabs}       
\usepackage{amsfonts}       
\usepackage{nicefrac}       
\usepackage{microtype}      
\usepackage{xcolor}         
\usepackage{epsfig}
\usepackage{graphicx}
\usepackage{amsmath}
\usepackage{amssymb}
\usepackage{multirow}
\usepackage{color, colortbl}
\usepackage{subcaption}
\usepackage{subfloat}
\usepackage{tabularx}
\usepackage[export]{adjustbox}
\usepackage{threeparttable}
\usepackage{pifont}
\usepackage[misc]{ifsym} 

\definecolor{COLOR_CSID}{HTML}{e0f5ff}
\definecolor{COLOR_NEAROOD}{HTML}{ffefe0}
\definecolor{COLOR_FAROOD}{HTML}{ffdebf}
\definecolor{COLOR_MEAN}{HTML}{f0f0f0}

\newcommand{\pp}{p}
\newcommand{\RR}{\mathbb{R}}
\newcommand{\FF}{\mathbf{F}}
\newcommand{\XX}{\mathbf{X}}
\newcommand{\YY}{\mathbf{Y}}

\usepackage{xspace}
\usepackage{comment}
\usepackage{bm}
\usepackage{wrapfig}
\usepackage[pagebackref=false, breaklinks=true,letterpaper=true, colorlinks,citecolor=citecolor, linkcolor=linkcolor, bookmarks=false]{hyperref}
\definecolor{citecolor}{HTML}{0071BC}
\definecolor{linkcolor}{HTML}{ED1C24}

\makeatletter
\renewcommand\paragraph{
  \@startsection{paragraph} 
  {4} 
  {\z@} 
  {.5em \@plus1ex \@minus.2ex} 
  {-1.5em} 
  {\normalfont\normalsize\bfseries} 
}
\DeclareRobustCommand\onedot{\futurelet\@let@token\@onedot}
\def\@onedot{\ifx\@let@token.\else.\null\fi\xspace}

\makeatother

\begin{document}
\sloppy 


\title{SFNet: Faster and Accurate Semantic Segmentation via Semantic Flow}

\author{Xiangtai Li$^{1,2}*$ \and
        Jiangning Zhang$^{3}$\and
        Yibo Yang$^{1,6}$ \and
        Guangliang Cheng$^{2 (\textrm{\Letter}) }$ \and
        Kuiyuan Yang$^{5}$ \and
        Yunhai Tong$^{1 (\textrm{\Letter}) }$ \and
        Dacheng Tao$^{4,6}$
}
        
\institute{
    * Work done at SenseTime Research, Beijing. \textrm{\Letter}: Corresponding Authors. \\
    Xiangtai Li (lxtpku@pku.edu.cn) \\ 
    Yunhai Tong (yhtong@pku.edu.cn) \\
    Guangliang Cheng (guangliangcheng2014@gmail.com) \\
    1 National Key Laboratory of General Artificial Intelligence, School of Intelligence Science and Technology, Peking University \\
    2 SenseTime Research ; 3 Zhejiang University \\
    4 School of Computer Science, Faculty of Engineering, The University of Sydney, Darlington, NSW 2008, Australia. \\
    5 Xiaomi Car; 6 JD Explore Academy
    }
\date{Received: date / Accepted: date}


\maketitle

\begin{abstract}
In this paper, we focus on exploring effective methods for faster and accurate semantic segmentation. A common practice to improve the performance is to attain high-resolution feature maps with strong semantic representation. Two strategies are widely used: atrous convolutions and feature pyramid fusion, while both are either computationally intensive or ineffective. Inspired by the Optical Flow for motion alignment between adjacent video frames, we propose a Flow Alignment Module (FAM) to learn \textit{Semantic Flow} between feature maps of adjacent levels and broadcast high-level features to high-resolution features effectively and efficiently. Furthermore, integrating our FAM to a standard feature pyramid structure exhibits superior performance over other real-time methods, even on lightweight backbone networks, such as ResNet-18 and DFNet. Then to further speed up the inference procedure, we also present a novel Gated Dual Flow Alignment Module to directly align high-resolution feature maps and low-resolution feature maps where we term the improved version network as SFNet-Lite. Extensive experiments are conducted on several challenging datasets, where results show the effectiveness of both SFNet and SFNet-Lite. In particular, when using Cityscapes test set, the SFNet-Lite series achieve 80.1 mIoU while running at 60 FPS using ResNet-18 backbone and 78.8 mIoU while running at 120 FPS using STDC backbone on RTX-3090.

Moreover, we unify four challenging driving datasets (\textit{i.e.}, Cityscapes, Mapillary, IDD, and BDD) into one large dataset, which we named Unified Driving Segmentation (UDS) dataset. It contains diverse domain and style information. We benchmark several representative works on UDS. Both SFNet and SFNet-Lite still achieve the best speed and accuracy trade-off on UDS, which serves as a strong baseline in such a challenging setting. The code and models are publicly available at \url{https://github.com/lxtGH/SFSegNets}.

\end{abstract}
\section{Introduction}

Semantic segmentation is a fundamental vision task that aims to classify every pixel in the images correctly. It involves many real-world applications, including \textit{auto-driving, robot navigation, and image editing}. The seminal work of Long~\emph{et. al.}~\cite{fcn} built a deep Fully Convolutional Network (FCN), which is mainly composed of convolutional layers to carve strong semantic representation. However, detailed object boundary information, which is also crucial to the performance, is usually missing due to the use of the down-sampling layers. 

To alleviate this problem, state-of-the-art methods~\cite{pspnet,psanet,DAnet,nvidia_seg_video} apply atrous convolutions~\cite{dilation} at the last several stages of their networks to yield feature maps with strong semantic representation while at the same time maintaining the high-resolution. Meanwhile, several state-of-the-art approaches~\cite{upernet,deeplabv3p,xiangtl_gff} adopt multiscale feature representation to enhance final segmentation results. Recently, several methods~\cite{cheng2021maskformer,wang2020maxDeeplab,SETR} adopt vision transformer architectures and model the semantic segmentation as a per-segment prediction problem. In particular, they achieve stronger performance for the long-tailed datasets, including ADE-20k~\cite{ADE20K} and COCO-stuff~\cite{coco_stuff} due to the stronger pre-trained models~\cite{liu2021swin} and query-based mask representation~\cite{detr}. 

Despite those methods achieving state-of-the-art results on various benchmarks, one fundamental problem is the real-time inference speed, particularly for high-resolution image inputs. Given that the FCN using ResNet-18~\cite{resnet} as the backbone network has a frame rate of 57.2 FPS for a $1024\times 2048$ image, after applying atrous convolutions~\cite{dilation} to the network as done in \cite{pspnet,psanet}, the modified network \textit{only has a frame rate of 8.7 FPS}. Moreover, under a single GTX 1080Ti GPU with no other ongoing programs, the previous state-of-the-art model PSPNet~\cite{pspnet} has a frame rate of only 1.6 FPS for $1024 \times 2048$ input images. Consequently, this is problematic for many advanced real-world applications, such as self-driving cars and robot navigation, which desperately demand real-time online data processing.

In order to not only maintain detailed resolution information but also get features that exhibit strong semantic representation, another direction is to build FPN-like~\cite{fpn,PanopticFPN,unet} models which leverage the lateral path to fuse feature maps in a top-down manner. In this way, the deep features of the last several layers strengthen the shallow features with high resolution, and therefore, the refined features are possible to satisfy the above two factors and are beneficial to the accuracy improvement. Such designs are mainly adopted by real-time semantic segmentation models. However, the accuracy of these methods~\cite{unet,segnet,swiftnet,peng2022pp} still needs improvement when compared to those networks that hold large feature maps in the last several stages. Is there a better solution for high accuracy and high-speed semantic segmentation? We suspect that the low accuracy problem arises from the ineffective propagation of semantics from deep layers to shallow layers, where the semantics are not well aligned across different stages.

\begin{figure}[!t]
	\centering
	\includegraphics[width=1.0\linewidth]{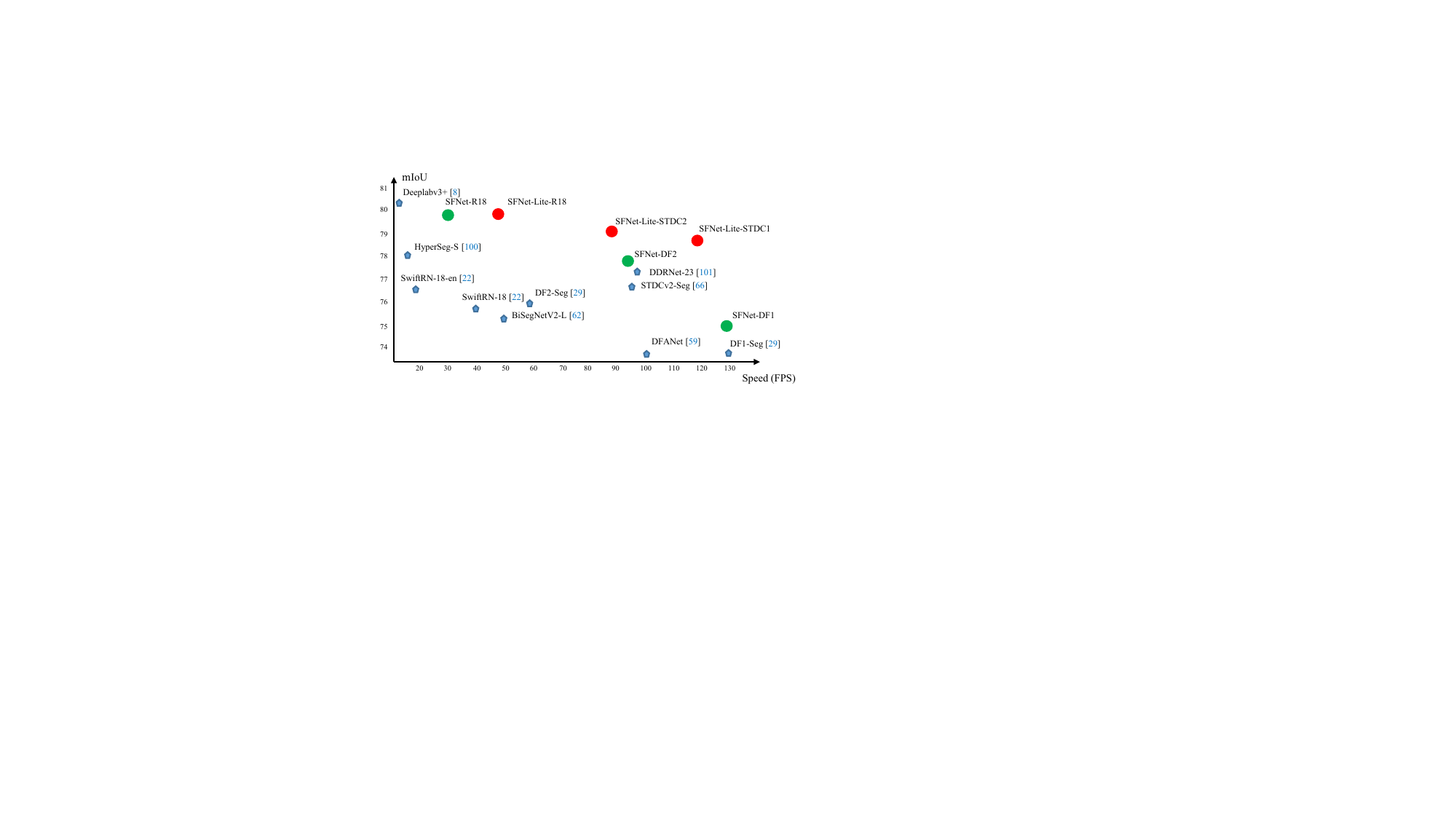}
	\caption{Inference speed versus mIoU performance on test set of Cityscapes. Previous models are marked as blue points, and our models are shown in red and green points which achieve the best speed/accuracy trade-off. Note that our methods with ResNet-18 as backbone even achieve comparable accuracy with all accurate models at much faster speed. SFNet methods are the \textcolor{green}{green nodes} while SFNet-Lite methods are the \textcolor{red}{red nodes}. }
	\label{fig:tesear_stoa_city}
\end{figure}

To mitigate this issue, we propose explicitly learning the \textbf{Semantic Flow} between two network layers of different resolutions. Semantic Flow is inspired by optical flow, which is widely used in video processing task~\cite{DFF} to represent the pattern of apparent motion of objects, surfaces, and edges in a visual scene caused by relative motion. \textit{In a flash of inspiration, we find the relationship between two feature maps of arbitrary resolutions from the same image can also be represented with the ``motion'' of every pixel from one feature map to the other one.}
In this case, once precise Semantic Flow is obtained, the network is able to propagate semantic features with minimal information loss. It should be noted that Semantic Flow is different from optical flow, since Semantic Flow takes feature maps from different levels as input and assesses the discrepancy within them to find a suitable flow field that will give a dynamic indication about how to align these two feature maps effectively.

\begin{figure}[!t]
	\centering
	\includegraphics[width=1.0\linewidth]{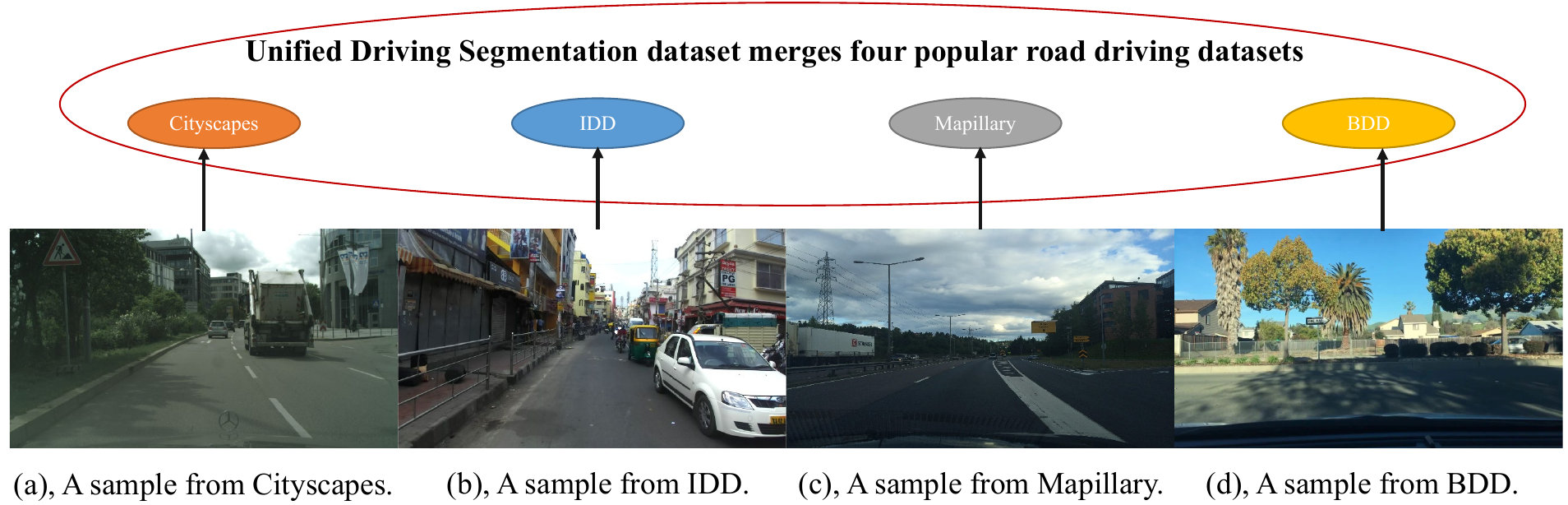}
	\caption{Illustration of the merged Unified Driving Segmentation (UDS) benchmark. It contains four datasets including Cityscapes~\cite{Cityscapes} (a), IDD~\cite{varma2019idd}(b), Mapillary~\cite{mapillary}(c) and BDD~\cite{yu2020bdd100k} (d). These datasets have \textit{various styles and texture information}, which make the merged UDS dataset more challenging.}
	\label{fig:tesear_uds}
\end{figure}

Based on the concept of Semantic Flow, we design a novel network module called Flow Alignment Module~(FAM) to utilize Semantic Flow in semantic segmentation. Feature maps after FAM are embodied with both rich semantics and abundant spatial information. Because FAM can effectively transmit semantic information from deep to shallow layers through elementary operations, it shows superior efficacy in improving accuracy and keeping superior efficiency. Moreover, FAM is end-to-end trainable and can be plugged into any backbone network to improve the results with a minor computational overhead. For simplicity, we call the networks that all incorporate FAM but have different backbones as \textbf{SFNet}. As depicted in Fig.~\ref{fig:tesear_stoa_city}, SFNets with different backbone networks outperform competitors by a large margin at the same speed. In particular, our method adopting ResNet-18 as backbone achieves \textbf{79.8\%} mIoU on the Cityscapes test server with a frame rate of \textbf{33 FPS}. When adopting DF2~\cite{DF-seg-net} as the backbone, our method achieves 77.8\% mIoU with 103 FPS and 74.5\% mIoU with 134 FPS when equipped with the DF1 backbone~\cite{DF-seg-net}. The results are shown in Fig.~\ref{fig:tesear_stoa_city} (\textcolor{green}{green node}). 

The original SFNet~\cite{SFnet} achieves satisfactory results on speed and accuracy trade-off, and several following works~\cite{huang2021fapn} generalize the idea of SFNet into other domains. However, the inference speed of SFNet still needs to be faster due to the multi-stage features involved. To speed up the SFNet and maintain accuracy at the same time, we propose a new version of SFNet, named SFNet-Lite. In particular, we design a new flow-aligned module named Gated Dual Flow Aligned Module (GD-FAM). Following FAM, GD-FAM takes two features as inputs and learns \textit{two semantic flows} to refine both high-resolution and low-resolution features simultaneously. Meanwhile, we also generate a shared gate map to control the flow warping processing before the final addition dynamically. The newly proposed GD-FAM can be appended at the end of SFNet backbone \textit{only once}, directly refining the highest and lowest resolution features. Such design avoids multiscale feature fusion and speeds up the SFNet by a large margin. We name our new version of SFNet as \textbf{SFNet-Lite}.
{Moreover, to keep the origin accuracy, we carry out extensive experiments on Cityscapes by introducing more balanced datasets training~\cite{nvidia_seg_video}. As a result, our SFNet-Lite with ResNet-18 backbone achieves \textbf{80.1} mIoU on Cityscapes test set but with the speed of \textbf{49 FPS} (\textbf{16 FPS} improvements with slightly better performance over original SFNet~\cite{SFnet}). Moreover, when adopting with STDCv1 backbone, our method can achieve \textbf{78.7 mIoU} while running with the speed of \textbf{120 FPS.} The results are shown in Fig.~\ref{fig:tesear_stoa_city} (\textcolor{red}{red node}).}

{Since various driving datasets~\cite{yu2020bdd100k,varma2019idd,Cityscapes} are from different domains, previous real-time semantic segmentation methods train different models on different datasets, which results in that the trained models are sensitive to trained domains and can not generalize well to unseen domain~\cite{choi2021robustnet}. Recently, M-Seg propose a mixed dataset for multi-dataset semantic to achive one model for multiple dataset training and test. Motivated by above, we verify whether our SFNet series can be more effective in a unified dataset benchmark. Firstly, we benchmark our SFNet and SFNet-Lite on various driving datasets~\cite{yu2020bdd100k,mapillary,varma2019idd} in the experiment part. Secondly, we creat a challenging benchmark by mixing four challenging driving datasets, including Cityscapes, Mapillary, BDD, and IDD. We term our merged dataset Unified Driving Segmentation (UDS).} As shown in Fig.~\ref{fig:tesear_uds}, our goal is to train \textbf{a unified model} to perform semantic segmentation on various scenes. To the best of our knowledge, UDS is the largest public semantic segmentation dataset for the driving scene. In particular, we extract the typical semantic class as defined by Cityscapes and BDD with 19 class labels and merge several classes in Mapillary. We further benchmark representative works on our UDS. {Our SFNet also achieves the best accuracy and speed trade-off, which \textit{indicates the generalization ability of semantic flow}. In particular, using DFNet~\cite{DF-seg-net} as the backbone, our SFNet and SFNet-Lite achieve \textit{7-9\% mIoU} improvements on UDS. This indicates that our proposed FAM and GD-FAM are more practical to multiple-dataset training.}

In summary, a preliminary version of this work was published in~\cite{SFnet}.
In this paper, we make the following significant extensions:
(1) We introduce a new flow alignment module (GD-FAM) to increase the speed of SFNet while maintaining the original performance. Experiments show that this new design consistently outperforms our previous module with higher inference efficiency. 
(2) We conduct more comprehensive ablation studies to verify the proposed method, including quantitative improvements over baselines and visualization analysis.
(3) We extend SFNet into Panoptic Segmentation, where we achieve 1.0\%-1.5\% PQ improvements over three strong baselines. 
{(4) We further benchmark SFNet and several recent representative methods on two more challenging datasets, including Mapillary~\cite{mapillary} and IDD~\cite{varma2019idd}. Our SFNet series significantly improve over different baselines and achieve the best speed and accuracy trade-off. In particular, we propose a new setting for training a unified real-time semantic segmentation model by merging existing driving datasets (UDS). Our SFNet series also achieve the best accuracy and speed trade-off, which can be a solid baseline for mixed driving segmentation. We further prove the effectiveness of SFNet and SFNet-Lite on transformer architecture on the ADE20k dataset. Moreover, aided by the RobustNet~\cite{choi2021robustnet}, we further show the effectiveness of SFNet on domain generalization setting.}
\section{Related Work}

\noindent
\textbf{Generic Semantic Segmentation.}
Current state-of-the-art methods on semantic segmentation are based on the FCN framework, which treats semantic segmentation as a dense pixel classification problem. Lots of methods focus on global context modeling with dilated backbone. Global average pooled features are concatenated into existing feature maps in ~\cite{parsenet}. In PSPNet~\cite{pspnet}, average pooled features of multiple window sizes, including global average pooling, are upsampled to the same size and concatenated together to enrich global information. The DeepLab variants \cite{deeplabv1,deeplabv3,deeplabv3p} propose atrous or dilated convolutions and atrous spatial pyramid pooling (ASPP) to increase the effective receptive field. DenseASPP~\cite{denseaspp} improves on \cite{deeplabv2} by densely connecting convolutional layers with different dilation rates to further increase the receptive field of the network. In addition to concatenating global information into feature maps, multiplying global information into feature maps also shows better performance~\cite{encodingnet,cbam,cgnl,dfn}. Moreover, several works adopt the self-attention design to encode the global information for the scene. Using non-local operator~\cite{Nonlocal}, impressive results are achieved in~\cite{ocnet,CoCurrentNet,DAnet}. CCNet~\cite{ccnet} models the long-range dependencies by considering its surrounding pixels on the criss-cross path via a recurrent way to save computation and memory cost. Meanwhile, several works~\cite{unet,upernet,PanopticFPN,li2021pointflow,eblnet_iccv} adopt encode-decoder architecture to learn the multi-level feature representation. RefineNet~\cite{refinenet} and DFN~\cite{dfn} adopted encoder-decoder structures that fuse information in low-level and high-level layers to make dense prediction results. Following such architecture design, GFFNet~\cite{xiangtl_gff}, CCLNet~\cite{ding2018context}, and G-SCNN~\cite{gated-scnn} use gates for feature fusion to avoid noise and feature redundancy. AlignSeg~\cite{huang2021alignseg} proposes to refine the multiscale features via bottom-up design. IFA~\cite{hu2022learning} proposes an implicit feature alignment function to refine the multiscale feature representation. In contrast, our method transmits semantic information top-down, focusing on real-time application. However, only some of these works can perform inference in real-time, which makes it hard to employ in practical applications.

\noindent
\textbf{Vision Transformer based Semantic Segmentation.} Recently, transformer-based approaches~\cite{VIT,liu2021swin,SETR,yuan2021polyphonicformer} replace the CNN backbones with vision transformers and achieve more robust results. Several works~\cite{SETR,liu2021swin,xie2021segformer,Segmenter} show that the vision transformer backbone leads to better results on long-tailed datasets due to the better feature representation and stronger pre-training on ImageNet classification. SETR~\cite{SETR} replaces the pixel level modeling with token-based modeling, while Segformer~\cite{xie2021segformer} proposes a new efficient backbone for segmentation. Moreover, several works~\cite{wang2020maxDeeplab,cheng2021maskformer,zhang2021knet} adopt Detection Transformer (DETR)~\cite{detr} to treat per-pixel prediction as a per-mask prediction. In particular, Maskformer~\cite{cheng2021maskformer} treats the pixel-level dense prediction as a set prediction problem. However, all of these works still can not perform inference in real-time due to the huge computation cost.

\noindent
\textbf{Fast Semantic Segmentation.} Fast (Real-time) semantic segmentation algorithms attract attention when demanding practical applications that need fast inference and response. Several works are designed for this setting. ICNet~\cite{ICnet} uses multiscale images as input and a cascade network to be more efficient. DFANet~\cite{dfanet} utilizes a light-weight backbone to speed up its network and proposes a cross-level feature aggregation to boost accuracy, while SwiftNet~\cite{swiftnet} uses lateral connections as the cost-effective solution to restore the prediction resolution while maintaining the speed. ICNet~\cite{ICnet} reduces the high-resolution features into different scales to speed up the inference time. ESPNets~\cite{ESPNet,ESPNetv2} save computation by decomposing standard convolution into point-wise convolution and spatial pyramid of atrous convolutions. BiSeNets~\cite{bisenet,yu2021bisenetv2} introduce spatial path and semantic path to reduce computation. Recently, several methods~\cite{fast_cell_search_seg,custom_search_seg,DF-seg-net} use AutoML techniques to search efficient architectures for scene parsing. Moreover, there are several works~\cite{STDCNet,si2019real} using multi-branch architecture to improve the real-time segmentation results.
However, these works result in poor segmentation results compared with those general methods on multiple benchmarks such as Cityscapes~\cite{Cityscapes} and Mapillary~\cite{mapillary}. Our previous work SFNet~\cite{SFnet} achieves high accuracy via learning semantic flow between multiscale features while running in real-time. However, its inference speed is still limited since more features are involved. Moreover, the capacity of multiscale features needs to be better explored via stronger data augmentation and pre-training. Thus, simultaneous achievement of high speed and high accuracy is still challenging and of great importance for real-time application purposes. 

\noindent
\textbf{Panoptic Segmentation.} Earlier works~\cite{kirillov2019panopticfpn,li2019attention,chen2020banet,porzi2019seamless,yang2019sognet} are proposed to model both stuff segmentation and thing segmentation in one model with different task heads. Detection-based methods~\cite{xiong2019upsnet,kirillov2019panopticfpn,qiao2021detectors,hou2020real} usually represent things with the box prediction, while several bottom-up models~\cite{cheng2020panoptic,axialDeeplab} perform grouping instance via pixel-level affinity or center heat maps from semantic segmentation results. The former introduces the complex process, while the latter suffers from performance drops in complex scenarios. Recently, several works~\cite{wang2020maxDeeplab,zhang2021knet,cheng2021maskformer} propose directly obtaining segmentation masks without box supervision. However, all of these works ignore the speed issue. In the experiment, we further show that our method can also lead to better panoptic segmentation results.

\noindent
\textbf{Lightweight Architecture Design.} Another critical research direction is to design more efficient backbones for the downstream tasks via various approaches~\cite{howard2017mobilenets,sandler2018mobilenetv2,shufflenetv2,STDCNet}. These methods focus on efficient representation learning with various network search approaches. Our work is orthogonal to those works, since we aim to design a lightweight and aligned segmentation head.

\noindent
\textbf{Multi-dataset Segmentation.} MSeg~\cite{MSeg_2020_CVPR} firstly proposes to merge most existing datasets in one unified taxonomy and train a unified segmentation model for variant scenes. {Meanwhile, several following works~\cite{zhou2022simple,li2022language} explore multi-dataset segmentation or detection. Compared with MSeg, our UDS dataset mainly focuses on the driving scene and has only 19 classes compared with more than 100 classes in MSeg. The input images are high-resolution and are used for auto-driving applications.}

{
\noindent
\textbf{Domain Generalization in Segmentation.} The goal domain generalization (DG)~\cite{wang2022generalizing}
methods assume that the model cannot access the target domain during training and aim to improve the generalization ability to perform well in an unseen target domain. DG is slightly different from multi-data segmentation. As for segmentation, several works~\cite{pan2018two,yue2019domain,kim2022pin,choi2021robustnet} adopt synthetic data such as GTAV for training and real dataset such as cityscapes for testing. Recently, RobustNet~\cite{choi2021robustnet} disentangles the domain-specific style and domain-invariant content encoded in higher-order statistics. Our method can also be applied in DG segmentation settings by combing RobustNet~\cite{choi2021robustnet}, where we also find significant improvements over various baselines.}

\begin{figure*}[!t]
	\centering
	\includegraphics[width=1.0\linewidth]{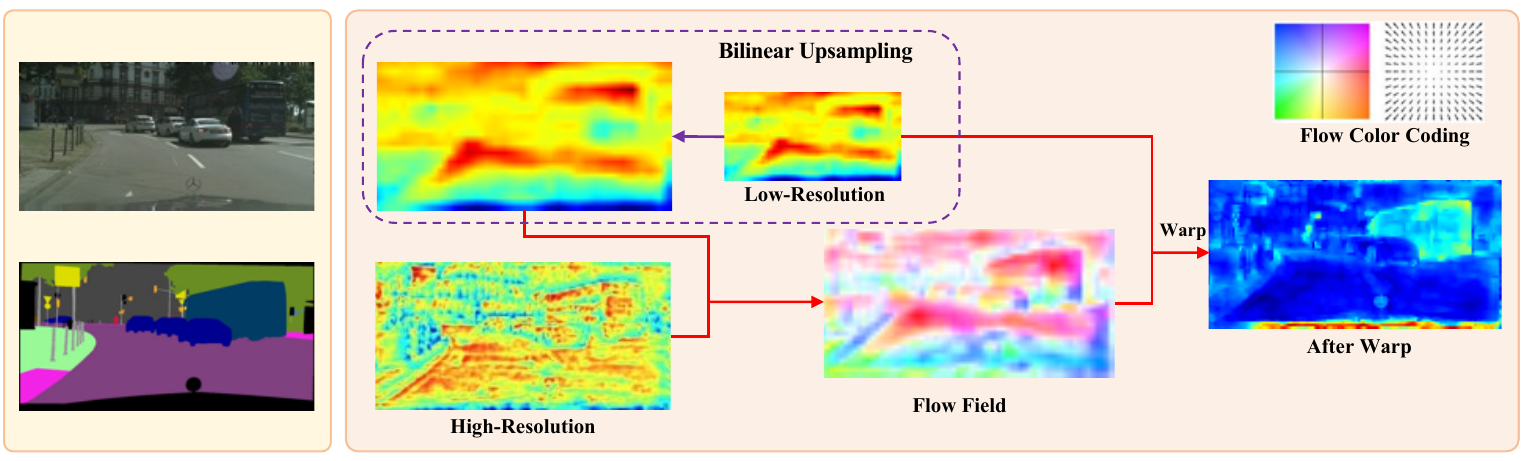}
	\caption{Visualization of feature maps and semantic flow field in FAM. Feature maps are visualized by averaging along the channel dimension. Larger values are denoted by hot colors and vice versa. We use the color code proposed in~\cite{flowvis} to visualize the Semantic Flow field. The orientation and magnitude of flow vectors are represented by hue and saturation, respectively. As shown in this figure, using our proposed semantic flow results in more structural feature representation. }
	\label{fig:issue}
\end{figure*}

\section{Method}

\noindent

In this section, we will first provide some preliminary knowledge about real-time semantic segmentation and introduce the misalignment problem therein. Then, we propose the Flow Alignment Module (FAM) to resolve the misalignment issue by learning Semantic Flow and warping top-layer feature maps accordingly. We also present the design of SFNet. Next, we introduce the proposed SFNet-Lite and the improved GD-FAM to speed up SFNet. Finally, we describe the building process of our UDS dataset and several improvement details for SFNet-Lite training.

\subsection{Preliminary}
The task of scene parsing is to map a RGB image $\XX \in \RR^{H\times W \times 3}$ to a semantic map $\YY \in \RR^{H\times W \times C}$ with the same spatial resolution $H\times W$, where $C$ is the number of predefined semantic categories. Following the setting of FPN~\cite{fpn}, the input image $\XX$ is firstly mapped to a set of feature maps $\{\FF_l\}_{l=2,...,5}$ from each network stage, where $\FF_l \in \RR^{H_l \times W_l \times C_l}$ is a $C_l$-dimensional feature map defined on a spatial grid $\Omega_l$ with size of $H_l \times W_l, H_l = \frac{H}{2^l}, W_l = \frac{W}{2^l}$.
The coarsest feature map $\FF_5$ comes from the deepest layer with the strongest semantics. FCN-32s directly predicts upon $\FF_5$ and achieves over-smoothed results without fine details. However, some improvements can be achieved by fusing predictions from lower levels~\cite{fcn}. FPN takes a step further to gradually fuse high-level feature maps with low-level feature maps in a top-down pathway through $2\times$ bilinear upsampling, which is originally proposed for object detection~\cite{fpn} and recently introduced for scene parsing~\cite{upernet,PanopticFPN}. 
The whole FPN framework highly relies on upsampling operator to upsample the spatially smaller but semantically stronger feature map to be larger in spatial size. However, the bilinear upsampling recovers the resolution of downsampled feature maps by interpolating a set of uniformly sampled positions (i.e., it can only handle one kind of fixed and predefined misalignment), while the misalignment between feature maps caused by residual connection, repeated downsampling and upsampling operations, is far more complex. Therefore, position correspondence between feature maps needs to be explicitly and dynamically established to resolve their actual misalignment.

\begin{figure*}[!t]
	\centering
	\includegraphics[width=1.0\linewidth]{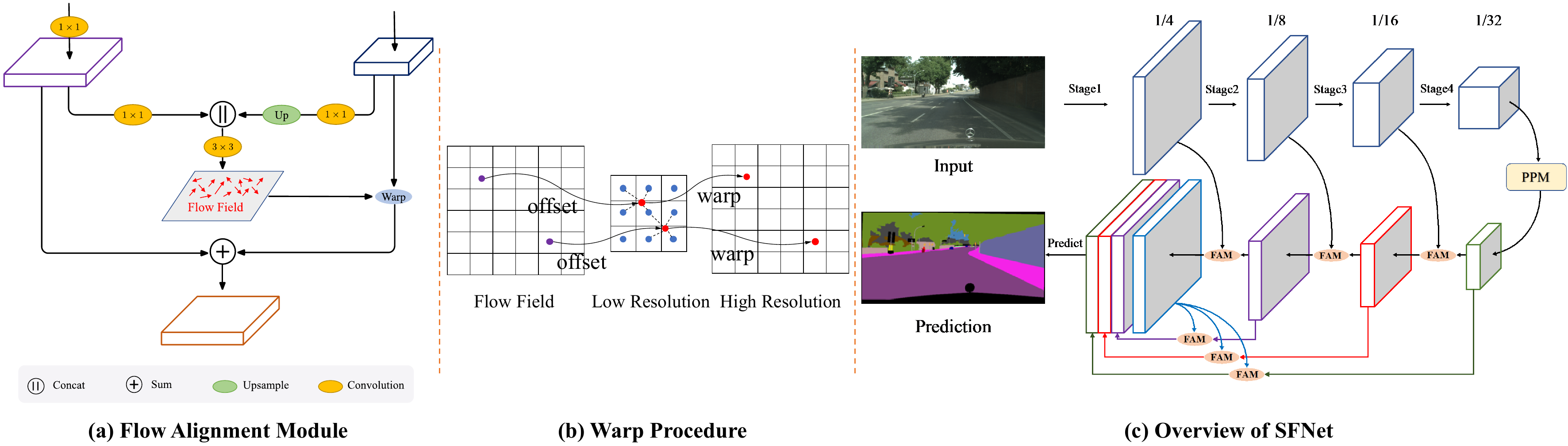}
	\caption{(a) The details of Flow Alignment Module. We combine the transformed high-resolution feature map and low-resolution feature map to generate the semantic flow field, which is utilized to warp the low-resolution feature map to a high-resolution feature map. (b) Warp procedure of Flow Alignment Module. The value of the high-resolution feature map is the bilinear interpolation of the neighboring pixels in the low-resolution feature map, where the neighborhoods are defined according to the learned semantic flow field. (c) Overview of our proposed SFNet. ResNet-18 backbone with four stages is used for exemplar illustration. FAM: Flow Alignment Module. PPM: Pyramid Pooling Module~\cite{pspnet}. Best view it in color and zoom in.
	}
	\label{fig:overview}
\end{figure*}

\subsection{Original Flow Alignment Module and SFNet}

\noindent \textbf{Design Motivation.}~For more flexible and dynamic alignment, we thoroughly investigate the idea of optical flow, which is very effective and flexible for aligning two adjacent video frame features in the video processing task~\cite{brox2004high,DFF}. The idea of optical flow motivates us to design a \emph{flow-based alignment module} (\textbf{FAM}) to align feature maps of two adjacent levels by predicting a flow field inside the network. We define such flow field as \emph{Semantic Flow}, which is generated between different levels in a feature pyramid.

\noindent \textbf{Module Details.}
FAM is constructed using the FPN framework, which involves compressing the feature map of each level into the same channel depth using two 1$\times$1 convolution layers before passing it on to the next level.
Given two adjacent feature maps $\FF_{l}$ and $\FF_{l-1}$ with the same channel number, we up-sample $\FF_{l}$ to the same size as $\FF_{l-1}$ via a bi-linear interpolation layer. Then, we concatenate them together and take the concatenated feature map as input for a subnetwork that contains two convolutional layers with the kernel size of $3\times 3$. The output of the subnetwork is the prediction of the semantic flow field $\Delta_{l-1} \in \RR^{H_{l-1} \times W_{l-1} \times 2}$. Mathematically, the aforementioned steps can be written as:
\begin{equation}
    \Delta_{l-1} = \text{conv}_l(\text{cat}(\FF_{l}, \FF_{l-1})),
\end{equation}
where $\text{cat}(\cdot)$ represents the concatenation operation and $\text{conv}_l(\cdot)$ is the $3\times 3$ convolutional layer.
Since our network adopts the strided convolutions, which could lead to very low resolution, for most cases, the respective field of the 3$ \times $3 convolution $\text{conv}_l$ is sufficient to cover most large objects in that feature map. Note that, we discard the correlation layer proposed in FlowNet-C~\cite{FlowNet}, where positional correspondence is calculated explicitly. Because there exists a huge semantic gap between higher-level layer and lower-level layer, explicit correspondence calculation on such features is difficult and tends to fail for offset prediction. Furthermore, including a correlation layer to address this issue would increase the computational cost substantially, which contradicts our objective of developing a fast and accurate network.

After having computed $\Delta_{l-1}$, each position $\pp_{l-1}$ on the spatial grid $\Omega_{l-1}$ is then mapped to a point $\pp_{l}$ on the upper level $l$ via a simple addition operation. Since there exists a resolution gap between features and flow field as shown in Figure~\ref{fig:overview}(b), the warped grid and its offset should be halved as Equation~\ref{mapping},
\begin{equation}
    \pp_{l} = \frac{\pp_{l-1}+\Delta_{l-1}(\pp_{l-1})}{2}. 
    \label{mapping}
\end{equation}
We then use the differentiable bi-linear sampling mechanism proposed in the spatial transformer networks~\cite{STN}, which linearly interpolates the values of the 4-neighbors (top-left, top-right, bottom-left, and bottom-right) of $\pp_{l}$ to approximate the final output of the FAM, denoted by $\widetilde\FF_l(\pp_{l-1})$. Mathematically,
\begin{equation}
    \widetilde\FF_l(\pp_{l-1}) = \FF_l(\pp_{l}) = \sum_{\pp \in \mathcal{N}(\pp_{l})} w_\pp \FF_{l}(\pp),
    \label{interpolate}
\end{equation}
where 
$\mathcal{N}(\pp_{l})$ represents neighbors of the warped points $\pp_l$ in $\FF_l$ and $w_\pp $ denotes the bi-linear kernel weights estimated by the distance of warped grid. This warping procedure may look similar to the convolution operation of the deformable kernels in deformable convolution network~(DCN)~\cite{deformable}. However, our method has a lot of noticeable difference from DCN. First, our predicted offset field incorporates both higher-level and lower-level features to \emph{align the positions} between high-level and low-level feature maps, while the offset field of DCN moves the positions of the kernels according to the predicted location offsets in order to \emph{possess larger and more adaptive respective fields}. Second, our module focuses on aligning features, while DCN works more like an attention mechanism that attends to the salient parts of the objects. More detailed comparison can be found in the experiment part.

On the whole, the proposed FAM module is light-weight and end-to-end trainable because it only contains one 3$\times$3 convolution layer and one parameter-free warping operation in total. Besides these merits, it can be plugged into networks multiple times with only a minor extra computation cost overhead. Figure~\ref{fig:overview}(a) gives the detailed settings of the proposed module, while Figure~\ref{fig:overview}(b) shows the warping process. Figure~\ref{fig:issue} visualizes the feature maps of two adjacent levels, their learned semantic flow and the finally warped feature map. As shown in Figure~\ref{fig:issue}, the warped feature is more structurally neat than the normal bi-linear upsampled feature and leads to more consistent representation of objects, such as the bus and car. 

Figure~\ref{fig:overview}(c) illustrates the whole network architecture, which contains a bottom-up pathway as the encoder and a top-down pathway as the decoder. While the encoder has a backbone network offering feature representations of different levels, the decoder can be seen as a FPN equipped with several FAMs.

\begin{figure*}[!h]
	\centering
	\includegraphics[width=1.0\linewidth]{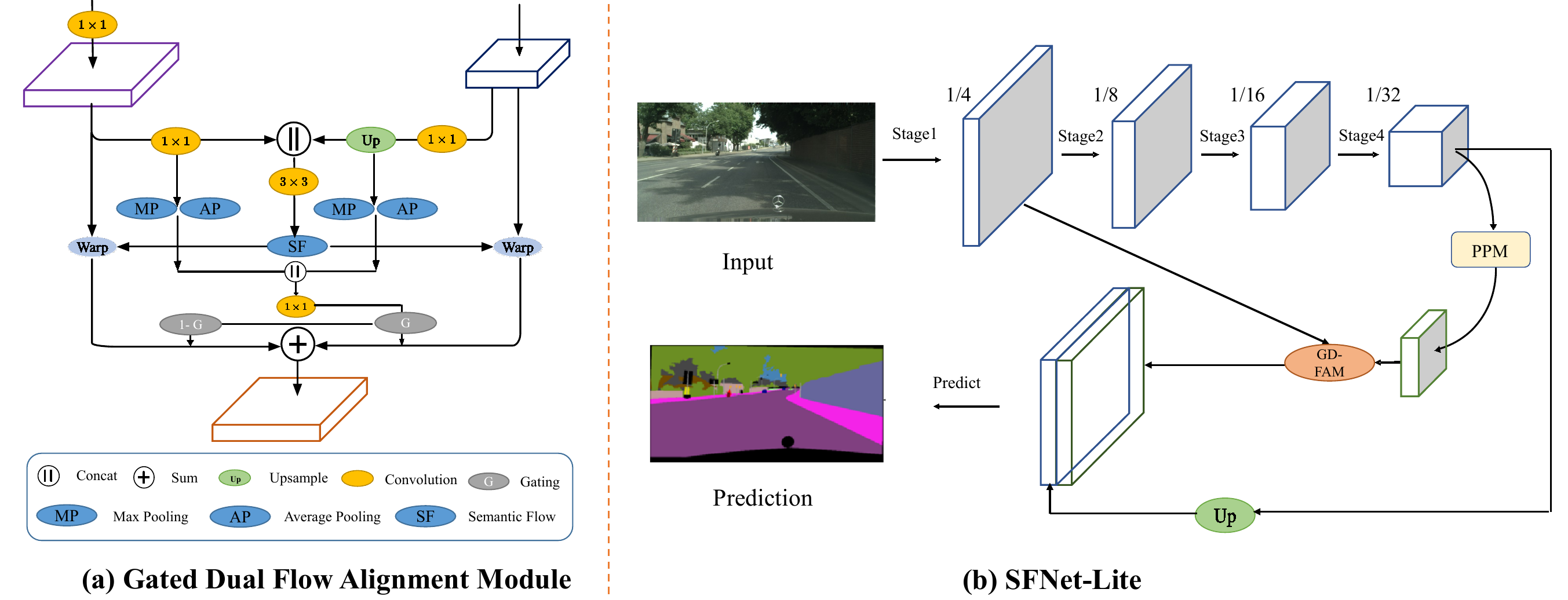}
	\caption{(a) The details of GD-FAM (Gated Dual Flow Alignment Module). We combine the transformed high-resolution feature map and low-resolution feature map to generate the two semantic flow fields and one shared gate map. The semantic flows are utilized to warp both the low-resolution feature map and the high-resolution feature map. The gate controls the fusion process. (b) Overview of our proposed SFNet-Lite. ResNet-18 backbone with four stages is used for exemplar illustration. GD-FAM: Gated Dual Flow Alignment Module. PPM: Pyramid Pooling Module~\cite{pspnet}. Best view it in color and zoom in.}
	\label{fig:sfnet_lite_method}
\end{figure*}

\noindent \textbf{Encoder Part.} We choose standard networks pre-trained on ImageNet~\cite{imagenet} for image classification as our backbone network by removing the last fully connected layer. Specifically, our experiments use and compare the ResNet series~\cite{resnet} and DF series~\cite{DF-seg-net}. All backbones consist of $4$ stages with residual blocks. To achieve both computational efficiency and larger receptive fields, we include a convolutional layer with a stride of 2 as the first layer in each stage, which downsamples the feature map. We additionally adopt the Pyramid Pooling Module (PPM)~\cite{pspnet} for its superior power to capture contextual information. In our setting, the output of PPM shares the same resolution as that of the last residual module. In this situation, we treat PPM and the last residual module together as the last stage for the upcoming FPN. Other modules like ASPP~\cite{deeplabv3} can also be plugged into our network, which is also experimentally ablated in the experiment part.

\noindent \textbf{Aligned FPN Decoder.} Our SFNet decoder takes feature maps from the encoder and uses the aligned feature pyramid for final scene parsing. By replacing normal bi-linear up-sampling with FAM in the top-down pathway of FPN~\cite{fpn}, $\{\FF_l\}_{l=2}^4$ is refined to $\{\widetilde{\FF}_l\}_{l=2}^4$, where top-level feature maps are aligned and fused into their bottom levels via element-wise addition and $l$ represents the range of feature pyramid level. For scene parsing, $\{\widetilde{\FF}_l\}_{l=2}^4 \cup \{\FF_5\}$ are up-sampled to the same resolution (\emph{i.e.}, 1/4 of the input image) and concatenated together for prediction. Considering there still exists misalignment during the previous step, we also replace these up-sampling operations with the proposed FAM. To be noted,  we only verify the effectiveness of such design in ablation studies. Our final models for the real-time application do not contain such a replacement for better speed and accuracy trade-off.

\subsection{Gated Dual Flow Alignment Module and SFNet-Lite}

\noindent
\textbf{Motivation.}
{Original SFNet adopts a multi-stage flow-based alignment process, it leads to a slower speed than several representative networks like BiSegNet~\cite{bisenet,ICnet}. Since the lightweight backbone design is not our main focus, we explore the \textbf{more compact decoder} with \textbf{only one} flow alignment module. Decreasing the number of FAM leads to inferior results (shown in experiment part, see Tab.~\ref{tab:city_ablation_FAM}(d)). To make up this gap, motivated by the recent success of gating design in segmentation~\cite{gated-scnn,xiangtl_gff}, we propose a new FAM variant named Gated Dual Flow Alignment Module (GD-FAM) to directly align and fuse both \textit{highest-resolution feature} and \textit{lowest-resolution feature}. Since there is only one aligment, which means less operators are involved, we can speed up the inference time.}

\noindent
\textbf{Gated Dual Flow Alignment Module.} As FAM, GD-FAM takes two features $\FF_4$ and $\FF_1$ as inputs and directly outputs a refined high resolution feature. We up-sample $\FF_{4}$ to the same size as $\FF_{1}$ via a bi-linear interpolation layer. Then, we concatenate them together and take the concatenated feature map as input for a subnetwork $conv_{F}$ that contains two convolutional layers with the kernel size of $3\times 3$. Such network directly outputs a new flow map $\Delta_{F} \in \RR^{H_{4} \times W_{4} \times 4}$. 
\begin{equation}
    \Delta_{F} = \text{conv}_F(\text{cat}(\FF_{4}, \FF_{1})).
\end{equation}
{We split such map $\Delta_{F}$ into $\Delta_{F1}$ and $\Delta_{F4}$ to jointly align both $\FF_{1}$ and $\FF_{4}$. Moreover, we propose to a shared gate map to highlight most important area on both aligned features. Our key insight is to make full use of high level semantic feature and let the low level feature as a supplement of high level feature. In particular, we adopt another subnetwork $conv_{g}$ to that contains one convolutional layer with the kernel size of $1 \times 1$ and one Sigmoid layer to generate such gate map. To highlight the most important regions of both features, we adopt max pooling ($\mathrm{Maxpool}$) and average pooling ($\mathrm{Avepool}$) over both features. Then we concatenate all four maps to generate such learnable gating maps.} This process is shown as following:
\begin{equation}
    \Delta_{G} = \text{conv}_g(\text{cat}(\mathrm{Avepool}(\FF_{4}, \FF_{1})). \mathrm{Maxpool}(\FF_{4}, \FF_{1}))),
\end{equation}
Then we adopt $\Delta_{G}$ to weight the aligned high semantic features and use inversion of $\Delta_{G}$ to weight the aligned low semantic features as fusion process. The key insights are two folds. Firstly, sharing the same gates can better highlight the most salient region. Secondly, adopting the subtracted gating supplies the missing details in low resolution feature. Such process is shown as following:
\begin{equation}
   \FF_{fuse}  =  \Delta_{G} Wrap(\Delta_{F1}, F1) + (1 - \Delta_{G}) Wrap(\Delta_{F4}, F4).
\end{equation}
where the Wrap process is the same as Equation~\ref{interpolate}. Our key insight is that a better fusion of both features can lead to more fine-grained feature representation: rich semantic and high resolution feature map. The entire process is shown in Figure~\ref{fig:sfnet_lite_method}(a). 

\noindent
\textbf{Lite Aligned Decoder.} The Lite Aligned Decoder is the simplified version of Aligned Decoder, which contains one GD-FAM and one PPM. As shown in  Figure~\ref{fig:sfnet_lite_method}(b), the final segmentation head takes the output of $\FF_{fuse}$ and upsampled deep features in last stage as inputs and outputs the final segmentation map via one $1\times1$ convolution over the combined inputs. Lite Aligned Decoder speeds up the Aligned Decoder via involving less multiscale features (only two scales). Avoiding shortcut design can also lead to faster speed when deploying the models on devices for practical usage. More results can be found in the experiment part.

\begin{table}[!t]\setlength{\tabcolsep}{6pt}
	\centering
			\caption{Speed comparison (FPS) on different devices for SFNet and SFNet-Lite. We adopt Resnet-18 as backbone. The FPS is measured by $1024 \times 2048$ input images.}
		\label{tab:sfnet_sfnet_lite_speed_compare}
	\begin{threeparttable}
		\scalebox{0.75}{
			\begin{tabular}{l c c c c}
				\toprule[0.2em]
				Device & 1080-TI & TTIAN-RTX & 3090-RTX & TITAN-RTX(TensorRT) \\  
				\toprule[0.2em]
				SFNet & 18.1 &  20.1 & 24.2 & 33.3  \\
				SFNet-Lite & 26.5 & 27.2 & 40.2 & 48.9  \\ 
				\bottomrule[0.1em]
			\end{tabular}
		}
	\end{threeparttable}
\end{table}

\noindent
\textbf{Speed Comparison Analysis.} { In Table~\ref{tab:sfnet_sfnet_lite_speed_compare}, we compare the speed of SFNet and SFNet-Lite on different devices. SFNet-Lite runs faster on various devices. In particular, when deploying both on TensorRT, the SFNet-Lite is \textit{even much faster} than SFNet since it involves less cross scale branches and leads to better optimization for acceleration. }

\subsection{The Unified Driving Segmentation Dataset}

\begin{table}[!t]\setlength{\tabcolsep}{6pt}
	\centering
	\caption{Dataset Information of our merged UDS dataset. We merged Mapillary labels into cityscapes label format. }
	\label{table:dataset_info}
	\begin{threeparttable}
		\scalebox{0.75}{
			\begin{tabular}{l c c c}
				\toprule[0.2em]
				Dataset Name & Train Images & Validation Images & Number of Class Labels\\  
				\toprule[0.2em]
				Cityscapes & 2,975 & 500 & 19 \\
				IDD &  6,993 & 3,000 & 19 \\ 
				Mapillary &  18,000 & 2,000 & 65  \\
				BDD & 7,000 & 1,000  & 19 \\
				\midrule
				UDS (ours) & 34,968 & 6,500 & 19 \\
				\bottomrule[0.1em]
			\end{tabular}
		}
	\end{threeparttable}
\end{table}

\noindent
\textbf{Motivation.} {Learning a unified driving-target segmentation model is useful since the environment may change a lot during the moving of self-driving cars. MSeg~\cite{MSeg_2020_CVPR} presents a more challenging setting while we only focus on high resolution out-door driving scene. Since the concepts of road scenes are limited, we only have small label space compared with M-Seg, which it has several common scenes (COCO~\cite{coco_dataset}, ADE20k~\cite{ADE20K}).}

{We verify the effectiveness of our SFNet series on new setting for feature alignment in various domains \textit{without} introducing domain aware learning~\cite{choi2021robustnet}. The goal of UDS is to provide more fair comparison on driving scene segmentation. To our knowledge, we are the first to benchmark such large-scale driving datasets using one model.}

\noindent
\textbf{Data Process and Results.}{
We merge four challenging datasets including Mapillary~\cite{mapillary}, Cityscapes~\cite{Cityscapes}, IDD~\cite{varma2019idd} and BDD~\cite{yu2020bdd100k}. Since Mapillary has 65 class labels, we merge several semantic labels into one label. The merging process follows the previous work~\cite{choi2021robustnet}.
We set other labels as ignore region. In this way, we keep the same label definition as Cityscapes and IDD. For IDD dataset, we use the same class definition as Cityscapes and BDD. For BDD and Cityscapes datasets, we keep the original setting. The merged dataset UDS totally has 34,968 images for training and 6,500 images for testing. The details of the UDS dataset are shown in Table~\ref{table:dataset_info}. Moreover, we find that several recent self-attention based methods~\cite{DAnet,OCRnet,EMANet} cannot perform well than previous method DeeplabV3+~\cite{deeplabv3p}. This implies a better generalized method is needed for this setting. We provide the code and model on the github pages. }

\noindent
\textbf{Discussion.}
Note that despite designing more balanced sampling methods or including domain generalization based method can improve the results on UDS, the goal of this work is only to verify the effectiveness of our SFNet and SFNet-Lite on this challenging setting.{ Both GD-GAM and FAM perform image feature level alignment, which are \textit{not sensitive} to the domain variations. Moreover, we also show the effectiveness of SFNet on domain generation settings using RobustNet~\cite{choi2021robustnet}. More details can be found in experiment part.}

\subsection{Improvement Details and Extension.}

\noindent \textbf{Improvement Details.} We use deeply supervised loss~\cite{pspnet} to supervise intermediate outputs of the decoder for easier optimization. In addition, following~\cite{bisenet}, online hard example mining~\cite{ohem} is also used by only training on the $10\%$ hardest pixels sorted by cross-entropy loss. During the inference, we only use the results from the main head. We also use uniform sampling methods to balance the rare class during training for all benchmarks. For the Cityscapes dataset, we also use the coarse boosting training tricks~\cite{nvidia_seg_video} to boost rare classes on Cityscapes. For backbone design, we also deploy the latest advanced backbone STDC~\cite{STDCNet} to speed up the inference speed on the device.

\noindent
\textbf{Extending SFNet into Panoptic Segmentation.} Panoptic Segmentation unifies both semantic segmentation and instance segmentation, which is a more challenging task. We also explore the proposed SFNet on such task with the proposed panoptic segmentation baseline K-Net~\cite{zhang2021knet}. K-Net is a state-of-the-art panoptic segmentation method where each thing and stuff is represented by kernels in its decoder head. In particular, we replace the backbone part of K-Net with our proposed SFNet backbone and aligned decoder. Then we train the modified model using the same setting as K-Net. 

\section{Experiment}

\subsection{Experiment Settings}

\noindent
\textbf{Overview.} We first review the dataset and training setting for SFNet. Then, we present the result comparison on five road-driving datasets, including the original SFNet and the newly proposed SFNet-lite. After that, we give detailed ablation studies and analysis on our SFNet. Finally, we present the generalization ability of SFNet on the Cityscapes Panoptic Segmentation dataset.

\noindent
\textbf{DataSets.} We mainly carry out experiments on the road driving datasets, including Cityscapes, Mapillary, IDD, BDD, and our proposed merged driving dataset. We also report panoptic segmentation results on the Cityscapes validation set. Cityscapes~\cite{Cityscapes} is a benchmark densely annotated for 19 categories of urban scenes, which contains 5,000 fine annotated images in total and is divided into 2,975, 500, and 1,525 images for training, validation, and testing, respectively. In addition, 20,000 coarse-labeled images are also provided to enrich the training data. Images are all with the same high resolution in the road driving scene, i.e., $1024 \times 2048$. Note that we use the fine-annotated dataset for ablation study and comparison with previous methods. We also use the coarse data to boost the final results of SFNet-Lite. Mapillary~\cite{mapillary} is a large-scale road-driving dataset, which is more challenging than Cityscapes since it contains more classes and various scenes. It contains 18,000 images for training and 2,000 images for validation. IDD~\cite{varma2019idd} is another road-driving dataset that mainly contains the India scene. It contains more images than Cityscapes.  It has 6,993 training images and 981 validation images. To our knowledge, we are \textit{the first} to benchmark the real-time segmentation models on Mapillary and IDD datasets. Another research group develops the BDD dataset, which mainly contains various scenes in American areas. It has 7,000  training images and 1,000 validation images.
\textit{All the datasets, including UDS dataset, are available online.}

\noindent
\textbf{Implementation Details.} We use PyTorch~\cite{pytorch} framework to carry out all the experiments. All networks are trained with the same setting, where stochastic gradient descent (SGD) with batch size of 16 is used as an optimizer, with a momentum of 0.9 and weight decay of 5e-4. All models are trained for 50K iterations with an initial learning rate of 0.01. As a common practice, the ``poly'' learning rate policy is adopted to decay the initial learning rate by multiplying $(1 -\frac{\text{iter}}{\text{total}\_\text{iter}})^{0.9}$ during training. Data augmentation contains random horizontal flip, random resizing with a scale range of $[0.75, ~2.0]$, and random cropping with crop size of $1024 \times 1024$ for Cityscapes, Mapillary, BDD, IDD, and UDS datasets. For quantitative evaluation, the mean of class-wise Intersection-Over-Union (mIoU) is used for an accurate comparison, and the number of Floating-point Operations Per Second (FLOPs) and Frames Per Second (FPS) are adopted for speed comparison. Moreover, to achieve a stronger baseline, we also adopt the class-balanced sampling strategy proposed in ~\cite{nvidia_seg_video}, which obtains stronger baselines. For the Cityscapes dataset, we also adopt coarse annotated data boosting methods to improve rare class segmentation quality. Our code and model are available for reference. Also note that several non-real segmentation methods in Mapillary, BDD, IDD, and USD datasets are implemented using our codebase and trained under the same setting.  

\noindent
\textbf{TensorRT Deployment Device.} 
The testing environment is TensorRT 8.2.0 with CUDA 11.2 on a single TITAN-RTX GPU. In addition, we re-implement the grid sampling operator by CUDA to be used together with TensorRT. The operator is provided by PyTorch and used in warping operations in the Flow Alignment Module. We report an average time of inferencing 100 images. Moreover, we also deploy our SFNet and SFNet-Lite on different devices, including 1080-TI and RTX-3090. We report the results in the next part.

\subsection{Main Results}

\noindent
\textbf{Results On Cityscapes test set.} We first report our SFNet on the Cityscapes dataset in Table~\ref{table:cityscapes_sota_speed_acc}. With ResNet-18 as the backbone, our method achieves 79.8\% mIoU and even reaches the performance of accurate models, which will be discussed next. Adopting STDC net as the backbone, our method achieves 79.8\% mIoU with full resolution inputs while running at 80 FPS. This suggests that our method can be benefited from a well-human-designed backbone. For the improved SFNet-Lite, our method can achieve even better results than the original SFNet while running faster using ResNet-18 as the backbone. For the STDC backbone, our method achieves much faster speed while maintaining similar accuracy. In particular, using STDC-v1, our method achieves 78.8\% mIoU while running at 120 FPS, a new state-of-the-art result on balancing speed and accuracy. This indicates the effectiveness of our proposed GD-FAM. 

\textbf{\textit{Note that for fair comparison, in Table~\ref{table:cityscapes_sota_speed_acc}, following previous works~\cite{STDCNet,yu2021bisenetv2}, we report the speed using Tensor-RT devices.}} For the results on the remaining datasets, we only report GPU average inference time. The Original SFNet with ResNet-18 achieves 78.9 \% mIoU, and we adopt uniform sampling, coarse boosting, and long-time training, which leads to an extra 0.9 \% gain on the test set. The details can be found in the following sections.

\begin{table}[!t]\setlength{\tabcolsep}{6pt}
	\centering
			\caption{Comparison on Cityscapes {\it test} set with state-of-the-art real-time models. For a fair comparison, the input size is also considered, and all models use single-scale inference. The FPS of our SFNet is evaluated on TensorRT following~\cite{STDCNet}.}
		\label{table:cityscapes_sota_speed_acc}
	\begin{threeparttable}
		\scalebox{0.75}{
			\begin{tabular}{l c c c c}
				\toprule[0.2em]
				Method  &   InputSize & mIoU ($\%$) & \#FPS & \#Params \\
				\toprule[0.2em]
				ESPNet~\cite{ESPNet} & $512 \times 1024$ & 60.3 & 132 & 0.4M \\
				ESPNetv2~\cite{ESPNetv2} &  $512 \times 1024$ & 62.1 & 80 & 0.8M \\
				ERFNet~\cite{ERFNet} & $512 \times 1024 $& 69.7 & 41.9 & - \\
				BiSeNet(ResNet-18)~\cite{bisenet} &  $768 \times 1536$ &  $74.6$ & 43 & 12.9M  \\
				BiSeNet(Xception-39)~\cite{bisenet} &  $768 \times 1536$ &  $68.4$ & 72 & 5.8M  \\ 
				BiSeNetv2(ResNet-18)~\cite{yu2021bisenetv2} &  $768 \times 1536$ & 75.3 &  47.3 & - \\
				BiSeNetv2(Xception-39)~\cite{yu2021bisenetv2} &  $768 \times 1536$ & 72.6 &  156 & - \\
				ICNet~\cite{ICnet} & $1024 \times 2048$ &  69.5 & 34 & 26.5M \\
				DF1-Seg~\cite{DF-seg-net} & $1024 \times 2048$ & 73.0 & 80 & 8.6M  \\
				DF2-Seg~\cite{DF-seg-net} & $1024 \times 2048$ & 74.8 & 55 & 18.9M  \\
				SwiftNet~\cite{swiftnet} & $1024 \times 2048$  & 75.5 & 39.9 & 11.8M \\
				SwiftNet-ens~\cite{swiftnet} & $1024 \times 2048$  & 76.5 & 18.4 & 24.7M \\
				DFANet~\cite{dfanet} & $1024 \times 1024$ & 71.3 & 100 & 7.8M \\
				CellNet~\cite{custom_search_seg} & $768 \times 1536$ & 70.5  & 108 & -\\
				STDC1-Seg75~\cite{STDCNet} & $768 \times 1536$ & 75.3 & 126.7 & 12.0M \\ 
                STDC2-Seg75~\cite{STDCNet} & $768 \times 1536$ & 76.8 & 97.0 & 16.1M\\
                HyperSeg-M~\cite{nirkin2021hyperseg}  & $512 \times 1024$ & 75.8 & 36.9 & 10.1M \\
                HyperSeg-S~\cite{nirkin2021hyperseg}  & $768 \times 1536$ &  78.1  & 16.1 &  10.2M \\
                DDRNet-23~\cite{hong2021deep} & $1024 \times 2048$  & 77.4 & 108.1 & 5.7M  \\
				\midrule
				SFNet(DF1) & $1024 \times 2048$ & ${74.5}$ & 134.5 & 9.0M \\
				SFNet(DF2) & $1024 \times 2048$ & ${77.8}$ & 103.1 & 19.6M \\
				SFNet(ResNet-18) & $1024 \times 2048$ & ${79.8}$ & 33.3 & 12.9M \\
				\midrule
				SFNet(STDC-1)  &  $1024 \times 2048$  &  78.1  & 97.1  & 9.1M   \\
				SFNet(STDC-2)  &  $1024 \times 2048$  &  79.8 & 79.9 & 13.1M  \\
				\midrule
				SFNet-Lite(ResNet-18) &  $1024 \times 2048$ & 80.1 &  48.9 &  12.3M \\
				SFNet-Lite(STDC-1) &  $1024 \times 2048$ &  78.8  &  119.1  &  9.7M \\
 				SFNet-Lite(STDC-2) &  $1024 \times 2048$ &  79.0  &  92.3  &  13.7M \\
				\bottomrule[0.1em]
			\end{tabular}
		}
	\end{threeparttable}
\end{table}

\begin{table}[!t]\setlength{\tabcolsep}{6pt}
	\centering
			\caption{Comparison on Mapillary {\it validation} set with state-of-the-art models. All the models are re-trained for a fair comparison and use single-scale inference with the same resolution input. The non-real-time models in the first sub-table use ResNet-50 as the backbone. The mIoU and FPS are measured {input image size with 1536 $\times$ 1536.} All the models are tested on a single TITAN-RTX.}
		\label{table:mapillary_sota_speed_acc}
	\begin{threeparttable}
		\scalebox{0.90}{
			\begin{tabular}{l c c c}
				\toprule[0.2em]
				Method   & mIoU ($\%$) & \#FPS & \#Params \\
				\toprule[0.2em]
				PSPNet~\cite{pspnet} & 42.4 &  4.8  & 31.1M \\
				Deeplabv3+~\cite{deeplabv3p} & 46.4 &  3.2 & 40.5M \\
				DANet~\cite{DAnet} & 42.9  &  2.0  & 48.1M \\
				OCRNet~\cite{OCRnet} & 46.6 & 3.8  &  39.0M \\
				EMANet~\cite{EMANet} & 47.5 &  4.2  & 34.8M \\
 				\midrule
				BiSeNet-V1(ResNet-18)~\cite{bisenet} &   43.2 & 24.3  & 12.9M  \\
				ICNet~\cite{ICnet}  & 42.8 &  48.2 & 26.5M \\
				DF1-Seg~\cite{DF-seg-net}  & 35.8 &  125.1  & 8.6M  \\
				DF2-Seg~\cite{DF-seg-net}  & 40.2 &  75.2  & 18.9M  \\
				STDC1~\cite{STDCNet}   & 41.9 &  34.5  & 12.0M \\ 
				STDC2~\cite{STDCNet}   & 43.5 &  29.0  & 16.1M \\ 
				\midrule
				SFNet(DF1)   & 41.4 &  102.2 & 9.0M \\
				SFNet(DF2)   & 45.6  &  57.8  & 19.6M \\
				SFNet(ResNet-18)   & 46.5 &  19.8  & 12.9M \\
				\midrule
				SFNet-Lite(ResNet-18) & 46.3 & 24.5 & 12.3M  \\
 				SFNet-Lite(STDC-2)  & 45.8 &   35.8 & 13.7M \\
				\midrule
				\bottomrule[0.1em]
			\end{tabular}
		}

	\end{threeparttable}
\end{table}

\begin{table}[!t]\setlength{\tabcolsep}{6pt}
	\centering
			\caption{Comparison on IDD {\it validation} set with state-of-the-art models.  For a fair comparison, all the models are re-trained and use single scale inference with the same resolution inputs ($ 1080 \times 1920 $ original size of IDD). }
		\label{table:idd_sota_speed_acc}
	\begin{threeparttable}
		\scalebox{0.90}{
			\begin{tabular}{l c c c}
				\toprule[0.2em]
				Method  & mIoU ($\%$) & \#FPS & \#Params \\
				\toprule[0.2em]
				PSPNet~\cite{pspnet} & 77.6 & 5.2 & 31.1M \\
				Deeplabv3+~\cite{deeplabv3p} & 78.9& 3.5 & 40.5M \\
				DANet~\cite{DAnet} & 76.6 & 3.2 & 48.1M \\
				OCRNet~\cite{OCRnet} & 78.1  & 4.2 & 39.0M\\
				EMANet~\cite{EMANet} & 77.2 & 4.4  & 34.8M\\
 				\midrule
				BiSeNet-V1(ResNet-18)~\cite{bisenet} & 74.4  & 24.5 & 12.9M  \\
				ICNet~\cite{ICnet} & 73.8 & 37.5 & 26.5M \\
				DF1-Seg~\cite{DF-seg-net} &  63.4 & 79.2 & 8.55M  \\
				DF2-Seg~\cite{DF-seg-net} & 67.9 & 50.8  & 18.88M  \\
				STDC1~\cite{STDCNet}   &  75.5 &  30.8 & 12.0M \\ 
				STDC2~\cite{STDCNet}   &  76.3 & 24.8 & 16.1M \\ 
				Bi-Align~\cite{bialign} & 73.9 &  30.2  & 19.2M \\
				\midrule
				SFNet(DF1) &  75.8 & 65.8 & 9.03M \\
				SFNet(DF2) &  76.3 & 37.4  & 19.63M \\
				SFNet(ResNet-18)  & 76.8 & 20.2  & 12.87M \\
				\midrule
				SFNet-Lite(ResNet-18) & 76.2 &  26.8  & 12.3M\\
 				SFNet-Lite(STDC-2)& 76.8  & 26.2  &  13.7M \\
				\bottomrule[0.1em]
			\end{tabular}
		}
	\end{threeparttable}
\end{table}

\begin{table}[!t]\setlength{\tabcolsep}{6pt}
	\centering
			\caption{Comparison on BDD {\it validation} set with state-of-the-art models. For a fair comparison, all the models are re-trained and use single scale inference with the same resolution inputs ($720 \times 1280$ original size of BDD).}
		\label{table:bdd_sota_speed_acc}
	\begin{threeparttable}
		\scalebox{0.90}{
		\begin{tabular}{l c c c}
				\toprule[0.2em]
				Method  & mIoU ($\%$) & \#FPS & \#Params \\
				\toprule[0.2em]
				PSPNet~\cite{pspnet} & 62.3 & 11.2  & 31.1M  \\
				Deeplabv3+~\cite{deeplabv3p} & 63.6 & 7.3 & 40.5M \\
				DANet~\cite{DAnet} & 62.8  & 6.6 & 48.1M\\
				OCRNet~\cite{OCRnet} & 60.1 & 7.1 & 39.0M \\
				EMANet~\cite{EMANet} & 61.4 & 9.6 & 34.8M \\
 				\midrule
				BiSeNet-V1(ResNet-18)~\cite{bisenet} & 53.8   &  45.1 & 12.9M  \\
				ICNet~\cite{ICnet} & 52.4 &  39.5 & 26.5M \\
				Bi-Align~\cite{bialign} & 53.4 & 42.1 & 19.2M \\
				DF1-Seg~\cite{DF-seg-net} & 42.5  &  82.3  & 8.6M  \\
				DF2-Seg~\cite{DF-seg-net} & 47.8  &  53.4 & 18.9M  \\
				STDC1~\cite{STDCNet}   &  52.1 &  45.8 & 12.0M \\ 
				STDC2~\cite{STDCNet}   &  53.8 &  33.0  & 16.1M \\ 
				\midrule
				SFNet(DF1) & 55.4  &  70.3 & 9.0M \\
				SFNet(DF2) &  60.2 &  47.3 & 19.6M \\
				SFNet(ResNet-18)  & 60.6 &  35.6 & 12.9M \\
				\midrule
				SFNet-Lite(ResNet-18) &  60.6 &  44.3  & 12.3M \\
 				SFNet-Lite(STDC-2) &  59.4  & 29.8  &  13.7M  \\
				\bottomrule[0.1em]
			\end{tabular}
		}
	\end{threeparttable}
\end{table}

\begin{table}[!t]\setlength{\tabcolsep}{6pt}
	\centering
			\caption{Comparison on UDS {\it testing} set with state-of-the-art models.  All the models are re-trained for fair comparison and use single scale inference with the same resolution inputs ($1024 \times 2048$, on both resized images and ground truth).}
		\label{table:uds_sota_speed_acc}
	\begin{threeparttable}
		\scalebox{0.90}{
		\begin{tabular}{l c c c}
				\toprule[0.2em]
				Method  & mIoU ($\%$) & \#FPS & \#Params \\
				\toprule[0.2em]
				PSPNet~\cite{pspnet} &  75.2 & 5.3 & 31.1M\\
				Deeplabv3+~\cite{deeplabv3p} & 78.0 & 3.7  & 40.5M \\
				DANet~\cite{DAnet}  & 75.8 & 3.0  & 48.1M \\
				OCRNet~\cite{OCRnet} & 77.0 & 4.2  & 39.0M\\
				EMANet~\cite{EMANet} & 76.8 & 4.4 & 34.8M \\
 				\midrule
				BiSeNet-V1(ResNet-18)~\cite{bisenet} & 73.8  &  24.5 & 12.9M  \\
				ICNet~\cite{ICnet} & 72.9 &  38.5 & 26.5M \\
				Bi-Align~\cite{bialign} & 73.9 & 30.1 & 19.2M \\
				DF1-Seg~\cite{DF-seg-net} & 62.6  & 75.2  & 8.6M  \\
				DF2-Seg~\cite{DF-seg-net} & 66.8 & 45.1 & 18.9M  \\
				STDC1~\cite{STDCNet}   &  74.0 & 30.2 & 12.0M \\ 
				STDC2~\cite{STDCNet}   &  75.2 &  24.5 & 16.1M \\ 
				\midrule
				SFNet(DF1) & 71.6  & 69.5   & 9.0M \\
				SFNet(DF2) &  75.5 &  37.5 & 19.6M \\
				SFNet(ResNet-18)  & 76.5 &  19.4 & 12.9M \\
				\midrule
				SFNet-Lite(ResNet-18)  & 75.3 &  24.5 & 12.3M\\
				SFNet-Lite(STDC-1) &  74.8 &  33.5 &  13.7M  \\
 				SFNet-Lite(STDC-2) &  75.6 & 30.2 &  13.7M  \\
				\bottomrule[0.1em]
			\end{tabular}
		}
	\end{threeparttable}
\end{table}

\noindent
\textbf{Results on Mapillary validation set.} In Table~\ref{table:mapillary_sota_speed_acc}, we report speed and accuracy results on a more challenging Mapillary dataset. Since this dataset contains huge resolution images and direct inference may raise the out-of-memory issue, we resize the short size of the image to 1,536 and crop the image and ground truth center following~\cite{nvidia_seg_video}. 

As shown in Table~\ref{table:mapillary_sota_speed_acc}, our methods also achieve the best speed and accuracy trade-off for various backbones. Even though the Deeplabv3+~\cite{deeplabv3p} and EMANet~\cite{EMANet} achieve higher accuracy, their speed cannot reach the real-time standard. In particular, for the DFNet-based backbone~\cite{DF-seg-net}, our SFNet achieves \textbf{almost 5-6\% mIoU} improvements. For SFNet-Lite, our methods also achieve considerable results while running faster. 

\noindent
\textbf{Results on IDD validation set.} In Table~\ref{table:idd_sota_speed_acc}, our methods achieve the best speed and accuracy trade-off. Compared with previous work STDCNet, our method achieves better accuracy and faster speed, as shown in the last row of Table~\ref{table:idd_sota_speed_acc}. For DFNet backbone, our methods also achieve nearly 12\% mIoU relative improvements. Such results indicate that the proposed FAM and GD-FAM accurately align the low-resolution feature into more accurate high-resolution and high-semantic feature maps.

\noindent
\textbf{Results on BDD validation set.} In Table~\ref{table:bdd_sota_speed_acc}, we further benchmark the representative works on BDD dataset. From that table, Deeplabv3+~\cite{deeplabv3p} achieves the top performance but with a much slower speed. 
Again, our methods, including both original SFNet and improved SFNet-Lite achieve the best speed and accuracy trade-off. For the recent state-of-the-art method STDCNet~\cite{STDCNet}, our SFNet-Lite achieves 5\% mIoU improvement while running slower. When adopting the ResNet-18 backbone, our SFNet-Lite achieves 60.6\% mIoU while running at 44.5 FPS without TensorRT acceleration.

\noindent
\textbf{Results on USD testing set.} Finally, we benchmark the recent works on the merged USD dataset in Table~\ref{table:uds_sota_speed_acc}. To fit the GPU memory, we resize both images and ground truth images to 1024 $\times$ 2048. From that table, we find Deeplabv3+~\cite{deeplabv3p} achieves top performance. Several self-attention-based models~\cite{EMANet,OCRnet,DAnet} achieve even worse results than previous Deeplabv3+ on such domain variant datasets. This shows that the USD dataset still leaves a huge room to improve. 

As shown in Table~\ref{table:uds_sota_speed_acc}, our methods using DFNet backbones achieve \textit{relatively 10\% mIoU improvements} over DF-Seg baselines. When equipped with the ResNet-18 backbone, our SFNet achieves 76.5\% mIoU while running at 20 FPS. When adopting the STDC-V2 backbone, our SFNet-Lite achieves the best speed and accuracy trade-off. 

\begin{table*}[!t]
    \footnotesize
	\centering
	\caption{Ablation studies on \textbf{SFNet architecture design} using Cityscapes validation set.}
	\label{tab:city_ablation_architecture}
    \scalebox{0.70}{\subfloat[Ablation study on baseline model.
    ]{
        \small
        \label{tab:effect_component}
	    \begin{tabular}{l c c c}
				\toprule[0.2em]
				Method &Stride & mIoU ($\%$) & $\Delta a (\%)$ \\
				\toprule[0.2em]
					FCN& 32 & 71.5  & -   \\
					Dilated FCN&8 & 72.6 & 1.1 $\uparrow$ \\
				\midrule
					+FPN &32 & 74.8 & 3.3 $\uparrow$ \\
					+FAM &32 & 77.2 & 5.7 $\uparrow$ \\ 
					+FPN + PPM &32 & 76.6 & 5.1 $\uparrow$ \\
					+FAM + PPM&32  & {78.3} & 7.2 $\uparrow$ \\
				\bottomrule[0.1em]
			    \end{tabular}
    }} \hfill
    \scalebox{0.70}{
    \subfloat[Ablation study on insertion position. 
    ]{
        \label{tab:dp_vs_jq}
	   \begin{tabular}{c c c c l l}
			\toprule[0.2em]
			Method & $\FF_3$ & $\FF_4$ & $\FF_5$ & mIoU(\%) & $\Delta a(\%) $ \\  
			\toprule[0.2em]
			FPN+PPM & - & - & - & 76.6 & -  \\ 
			& \checkmark & - & - & 76.9 & 0.3 $\uparrow$ \\
			& - & \checkmark & - & 77.0 & 0.4 $\uparrow$ \\
			& - & - & \checkmark & 77.5 & 0.9 $\uparrow$ \\
			\midrule
			\midrule
			& - &\checkmark &\checkmark & 77.8 & 1.2 $\uparrow$\\
			&\checkmark &\checkmark &\checkmark &78.3& 1.7 $\uparrow$\\
			\hline
		\end{tabular}
    }} \hfill
    \scalebox{0.65}{
    \subfloat[Ablation study on context module.]{
        \label{tab:aligned_decoder}
        \begin{tabular}{l l l l }
				\toprule[0.2em]
				Method & mIoU(\%) & $\Delta a(\%) $ & \#GFLOPs \\  
				\toprule[0.2em]
				FAM & 76.4  & -  & - \\
				+PPM~\cite{pspnet} & 78.3 & 1.9$\uparrow$ & 123.5 \\
				+NL~\cite{Nonlocal} & 76.8 & 0.4$\uparrow$ & 148.0 \\
				+ASPP~\cite{deeplabv3} & 77.6 & 1.2$\uparrow$ & 138.6 \\
				+DenseASPP~\cite{denseaspp} & 77.5 & 1.1$\uparrow$ & 141.5 \\
				\hline
			\end{tabular}
    }} \hfill
\end{table*}

\subsection{Ablation Studies}

\noindent
\textbf{Effectiveness of FAM and GD-FAM.} Table~\ref{tab:city_ablation_architecture}(a) reports the comparison results against baselines on the validation set of Cityscapes~\cite{Cityscapes}, where ResNet-18~\cite{resnet} serves as the backbone. Compared with the naive FCN, dilated FCN improves mIoU by 1.1\%. By appending the FPN decoder to the naive FCN, we get 74.8\% mIoU by an improvement of 3.2\%. By replacing bilinear upsampling with the proposed FAM, mIoU is boosted to 77.2\%, which improves the naive FCN and FPN decoder by 5.7\% and 2.4\%, respectively. Finally, we append PPM (Pyramid Pooling Module)~\cite{pspnet} to capture global contextual information, which achieves the best mIoU of 78.7 \% together with FAM. Meanwhile, FAM is complementary to PPM by observing FAM improves PPM from 76.6\% to 78.7\%. In Table~\ref{tab:city_ablation_gd_fam}(a), we compare the effectiveness of GD-FAM and FAM. As shown in that table, our new proposed GD-FAM has better performance (0.4\%) while running faster than the original FAM under the same settings. 

\noindent
\textbf{Positions to insert FAM or GD-FAM:} We insert FAM to different stage positions in the FPN decoder and report the results in Table~\ref{tab:city_ablation_architecture}(b). From the first three rows, FAM improves all stages and gets the greatest improvement at the last stage, demonstrating that misalignment exists in all stages of FPN and is more severe in coarse layers. This is consistent with the fact that coarse layers contain stronger semantics but with lower resolution and can greatly boost segmentation performance when they are appropriately upsampled to high resolution. The best result is achieved by adding FAM to all stages in the last row. For GD-FAM, we aim to align the high-resolution features and low-resolution directly. We choose to align $F_{3}$ and the output of PPM by default. 

\noindent
\textbf{Ablation study on network architecture design: } \label{sec:ablation}
Considering current state-of-the-art contextual modules are used as heads on dilated backbone networks~\cite{deeplabv3,denseaspp}, we further try different contextual heads in our methods where the coarse feature map is used for contextual modeling. Table~\ref{tab:city_ablation_architecture}(c) reports the comparison results, where PPM~\cite{pspnet} delivers the best result, while the more recently proposed methods such as non-Local-based heads~\cite{Nonlocal} perform worse. Therefore, we choose PPM as our contextual head due to its better performance with lower computational cost.

\begin{table*}[!t]
    \footnotesize
	\centering
		\caption{Ablation results on \textbf{FAM design} using Cityscapes validation set.}
	\label{tab:city_ablation_FAM}
    \scalebox{0.70}{\subfloat[Ablation study on Upsampling operation in FAM.
    ]{
        \small
	    \begin{tabular}{l c}
				\toprule[0.2em]
				Method  & mIoU ($\%$)  \\
				\toprule[0.2em]
					bilinear upsampling & 78.3 \\
					deconvolution & 77.9 \\ 
					nearest neighbor & 78.2 \\
 				\bottomrule[0.1em]
			    \end{tabular}
    }} \hfill
    \scalebox{0.70}{
    \subfloat[Ablation study on kernel size $k$ in FAM where 3 FAMs are involved. 
    ]{
	    \begin{tabular}{l c c}
				\toprule[0.2em]
				Method  & mIoU ($\%$) & Gflops\\
				\toprule[0.2em]
				    $k=1$ & 77.8 & 120.4 \\
				    $k=3$ & 78.3 & 123.5 \\
				    $k=5$ & 78.1 & 131.6 \\
                    $k=7$ & 78.0 & 140.5 \\
				\bottomrule[0.1em]
			    \end{tabular}
    }} \hfill
    \scalebox{0.65}{
    \subfloat[Ablation with FlowNet-C~\cite{FlowNet} in FAM.]{
	    \begin{tabular}{l c c c}
				\toprule[0.2em]
				Method  & mIoU ($\%$) & $\Delta a(\%) $ \\
				\toprule[0.2em]
					FPN +PPM  & 76.6  & -   \\
				\midrule
				    correlation~\cite{FlowNet} & 77.2 & 0.6 $\uparrow$ \\
					Ours  & 77.5 & 0.9 $\uparrow$ \\
				\bottomrule[0.1em]
			    \end{tabular}
    }} \hfill
    \scalebox{0.70}{
    \subfloat[Comparison with DCN~\cite{deformable}.]{
	     \begin{tabular}{c c c c l l}
			\toprule[0.2em]
			Method & $\FF_3$ & $\FF_4$ & $\FF_5$ & mIoU(\%) & $\Delta a(\%) $ \\  
			\toprule[0.2em]
			FPN +PPM & -  & - & - &  76.6 & - \\
			DCN & - & - &\checkmark & 76.9 & 0.3 $\uparrow$ \\
			Ours & - &  -&\checkmark & 77.5 & 0.9 $\uparrow$ \\
			\midrule
			\midrule
		    DCN	&\checkmark & \checkmark & \checkmark & 77.2 & 0.6 $\uparrow$\\
			Ours &\checkmark &\checkmark & \checkmark & 78.3 & 1.7 $\uparrow$\\
			\hline
		\end{tabular}
    }} \hfill
\end{table*}

\begin{table*}[!t]
    \footnotesize
	\centering
	\caption{Ablation experiment results on SFNet-Lite and \textbf{GD-FAM design} using Cityscapes validation set. DF: Dual Flow. Attn: Attention. G: Gate. US: Uniform Sampling. LT: Long Training. CB: Coarse Boosting.}
    \label{tab:city_ablation_gd_fam}
    \scalebox{0.70}{\subfloat[Effectiveness of GD-FAM. FPS is measuerd with 1024 $\times$ 2048 input. 
    ]{
	    \begin{tabular}{l  c  c}
				\toprule[0.2em]
				Method & mIoU ($\%$) & FPS \\
				\toprule[0.2em]
					FCN  & 71.5 &  50.3  \\
					+FPN + PPM (baseline) & 76.6  &  40.3 \\
					+ FAM + PPM   & 78.3  & 19.4 \\
					+ one GD-FAM + PPM & 78.3  & 24.5 \\
				\bottomrule[0.1em]
			    \end{tabular}
    }} \hfill
    \scalebox{0.70}{
    \subfloat[Ablation study on components in GD-FAM.
    ]{
	   \begin{tabular}{c c c c l }
			\toprule[0.2em]
			Method & DF & Atten & G & mIoU(\%)  \\  
			\toprule[0.2em]
			FPN+PPM & - & - & - & 76.6   \\ 
			& \checkmark & - & - & 77.8 \\
			& \checkmark & \checkmark &  - & 78.0  \\
			& \checkmark & \checkmark & \checkmark & 78.3 \\
			\hline
		\end{tabular}
    }} \hfill
    \scalebox{0.65}{
    \subfloat[Ablations study on Improving Tricks.]{
        \begin{tabular}{c c c c l }
			\toprule[0.2em]
			Method & US & LT & CB & mIoU(\%)  \\  
			\toprule[0.2em]
			SFNet-Lite &  - &  - & - & 78.3  \\ 
			& \checkmark & - &  - & 78.6 \\
			& \checkmark & \checkmark & - & 79.0  \\
			& \checkmark & \checkmark & \checkmark & 79.7 \\
			\hline
		\end{tabular}
    }} \hfill
\end{table*}

\begin{table}[!t]\setlength{\tabcolsep}{6pt}
	\centering
	\caption{Generalization on Various Backbone. For SFNet series, the baseline models are without FAM or GD-FAM. Note, GD-FAM is only used once. The GFlops are calculated with $1024 \times 2048$ input.}
		\label{table:different_backbone}
	\begin{threeparttable}
		\scalebox{0.90}{
			\begin{tabular}{l c c c c}
				\toprule[0.2em]
				Backbone & mIoU(\%) & $\Delta a(\%) $ & \#GFLOPs & $\Delta b(\%)$  \\  
				\toprule[0.2em]
				ResNet-50~\cite{resnet}& 76.8  & - & 332.6 & - \\
				w/ FAM & 79.2 & 2.4 $\uparrow$& 337.1 & +4.5 \\ 
				ResNet-101~\cite{resnet}& 77.6 & - & 412.7& \\
				w/ FAM  & 79.8 & 2.2$\uparrow$ & 417.5 & +4.8 \\
				w/ GD-FAM & 80.2 & 2.6$\uparrow$ & 415.3 & +2.6 \\
				\midrule
				ShuffleNetv2~\cite{shufflenetv2} & 69.8 & - & 17.8 & - \\
				w/ FAM  & 72.1 & 2.3 $\uparrow$& 18.1 & +0.3 \\
				DF1~\cite{DF-seg-net} & 72.1 & - & 18.6 & - \\
				w/ FAM  & 74.3 & 2.2 $\uparrow$ & 18.7 & +0.1 \\
				DF2~\cite{DF-seg-net} & 73.2 & - & 48.2 & -\\
				w/ FAM  & 75.8 & 2.6 $\uparrow$ & 48.5 & +0.3 \\
				\midrule
				STDC-Net-v1~\cite{STDCNet} & 75.0 & - & 58.2 & -\\
				w/ FAM & 76.7 & 1.7$\uparrow$  & 59.8 & +1.6 \\
				w/ GD-FAM & 76.5 &  1.5$\uparrow$ & 59.0 & +0.8 \\
				STDC-Net-v2~\cite{STDCNet} & 75.6 & - & 85.0 & - \\
				w/ FAM & 77.4 & 1.8$\uparrow$ & 86.3 & +1.3 \\
				w/ GD-FAM & 77.5  & 1.9$\uparrow$ & 86.2 & +1.2 \\
				\bottomrule[0.1em]
			\end{tabular}
		}
	\end{threeparttable}
\end{table}

\noindent
\textbf{Ablation on FAM design.}  We first explore the effect of upsampling in FAM in Table~\ref{tab:city_ablation_FAM}(a). Replacing the bilinear upsampling with deconvolution and nearest neighbor upsampling achieves 77.9\% mIoU and 78.2\% mIoU, respectively, which are similar to the 78.3\% mIoU achieved by bilinear upsampling. We also try the various kernel sizes in Table~\ref{tab:city_ablation_FAM}(b). A larger kernel size of $5\times5$ is also tried, which results in a similar result (78.2\%) but introduces more computation cost. In Table~\ref{tab:city_ablation_FAM}(c), replacing FlowNet-S with correlation in FlowNet-C also leads to slightly worse results (77.2\%) but increases the inference time. The results show that it is enough to use lightweight FlowNet-S for aligning feature maps in FPN. In Table~\ref{tab:city_ablation_FAM}(d), we compare our results with DCN~\cite{deformable}. We apply DCN on the concatenated feature map of the bilinear upsampled feature map and the feature map of the next level. We first insert one DCN in higher layers $\FF_{5}$ where our FAM is better than it. After applying DCN to all layers, the performance gap is much larger. This indicates that our method can also align low-level edges for better boundaries and edges in lower layers, which will be shown in the visualization part.

\begin{figure*}[!t]
	\centering
	\includegraphics[width=1.0\linewidth]{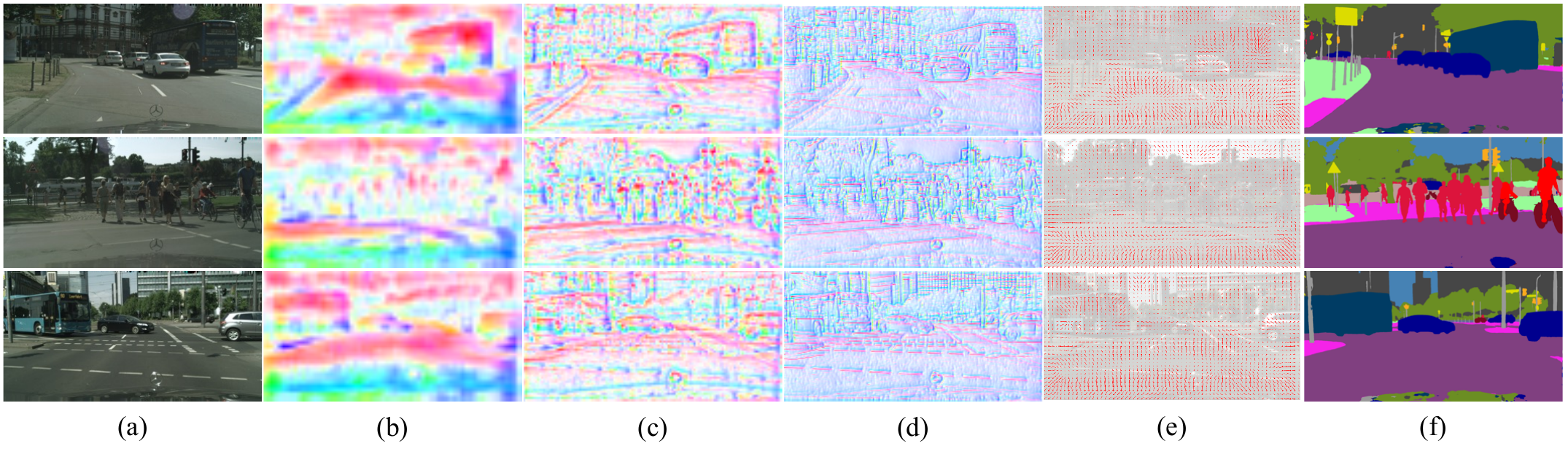}
	\caption{Visualization of the learned semantic flow fields. Column (a) lists three exemplary images. Column (b)-(d) show the semantic flow of the three FAMs in ascending order of resolution during the decoding process, following the same color coding of Figure~\ref{fig:issue}. Column (e) is the arrowhead visualization of flow fields in column (d). Column (f) contains the segmentation results.}
	\label{fig:vis_flowfield}
\end{figure*}

\noindent
\textbf{Ablation GD-FAM design.} In Table~\ref{tab:city_ablation_gd_fam}(b), we explore the effect of each component in GD-FAM. In particular, adding Dual Flow (DF) design boosts about 1.2\% improvement. Using Attention to generate gates rather than using convolution leads to 0.2\% improvement. Finally, using the shared gate design also improves the strong baseline by 0.3\%.

\noindent
\textbf{Ablation on Improving Details.} In Table~\ref{tab:city_ablation_gd_fam}(c), we explore the training tricks, including Uniform Sampling (US), Long Training (LT) and Coarse Boosting (CB). Performing US leads to 0.3\% improvements on our SFNet-Lite. Using LT (1000 epochs training) rather than short training (300 epochs training) results in another 0.4\% mIoU improvement. Finally, adopting coarse data boosts on several rare classes leads to another 0.7\% improvement.

\noindent
\textbf{generalization on Various Backbones.}  We further carry out experiments with different backbone networks, including both deep and light-weight networks, where the FPN decoder with PPM head is used as a strong baseline in Table~\ref{table:different_backbone}. For heavy networks, we choose ResNet-50 and ResNet-101~\cite{resnet} to extract representation. For light-weight networks, ShuffleNetv2~\cite{shufflenetv2}, DF1/DF2~\cite{DF-seg-net} and STDC-Net~\cite{STDCNet} are employed. FAM significantly achieves better mIoU on all backbones with slightly extra computational cost. Both GD-FAM and FAM improve the results of different backbones significantly with little extra computation cost.

\begin{figure}[!t]
	\centering
	\includegraphics[width=1.0\linewidth]{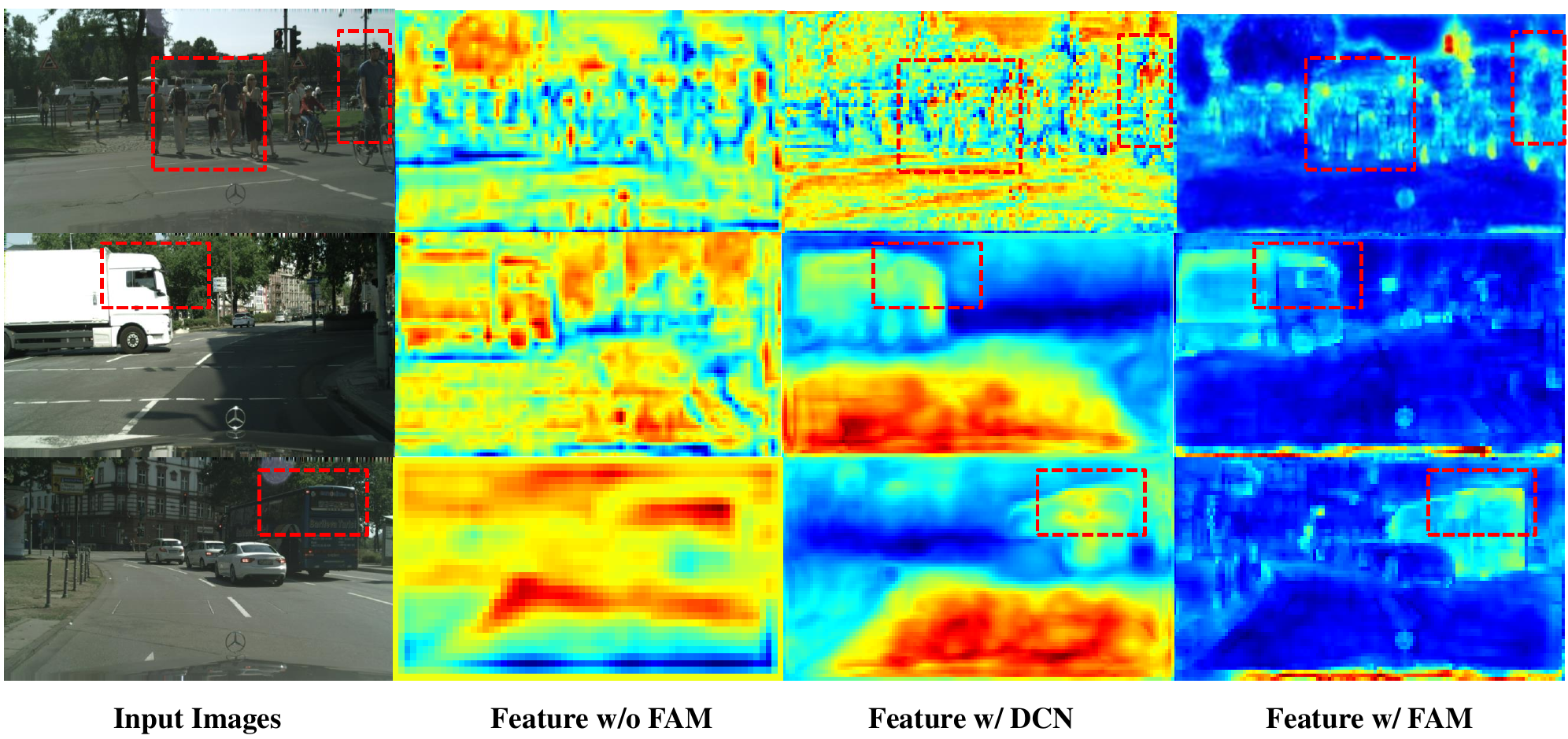}
	\caption{
	Visualization of the aligned feature. Compared with DCN, our module outputs more structural feature representation.}
	\label{fig:vis_align_fea}
\end{figure}

\noindent
\textbf{Aligned feature representation:} In this part, we present more visualization on aligned feature representation as shown in Figure~\ref{fig:vis_align_fea}. We visualize the upsampled feature in the final stage of ResNet-18. It shows that compared with DCN~\cite{deformable}, our FAM feature is more structural and has much more precise object boundaries, which is consistent with the results in Table~\ref{tab:city_ablation_FAM}(d). That indicates that FAM is \textbf{not} an attention effect on a feature similar to DCN, but aligns the feature towards a more precise shape than in red boxes.

\begin{table*}[t!]
\centering 
\small
\caption{Quantitative per-category comparison results on Cityscapes validation set, where ResNet-101 backbone with the FPN decoder and PPM head serves as the strong baseline. Obviously, Both FAM and GD-FAM boost the performance of almost all the categories.}
\label{tab:cityscapes_results_detail_val}
\addtolength{\tabcolsep}{0pt}
\resizebox{\textwidth}{!}{
			\begin{tabular}{ l | c c c c c c c c c c c c c c c c c c c | c}
		   	\toprule[0.2em]
			Method & road & swalk & build & wall & fence & pole & tlight & sign & veg. & terrain & sky & person & rider & car & truck & bus & train & mbike & bike & mIoU \\
			\toprule[0.2em]
			BaseLine & 98.1 & 84.9 & 92.6 & 54.8 & 62.2 & 66.0 & 72.8 & 80.8 & 92.4 & 60.6 & 94.8 & 83.1 & 66.0 & 94.9 & 65.9 & 83.9 & 70.5 & 66.0 & 78.9 & 77.6 \\ 
	        \hline
			with FAM  & 98.3 & 85.9 & 93.2 & 62.2 & 67.2 & 67.3 & 73.2 & 81.1 & 92.8 & 60.5 & 95.6 & 83.2 & 65.0 & 95.7 & 84.1 & 89.6 & 75.1 & 67.7 & 78.8 & 79.8 \\
			with GD-FAM  &  98.3 & 86.7 & 93.5 & 63.4 & 67.1 & 68.2 & 73.5 & 81.9 & 92.7 & 64.4 & 95.4 & 84.2 & 68.4 & 95.4 & 85.7 &  91.2 & 83.2 & 67.7 & 79.3 & 81.0 \\
		    \bottomrule[0.1em]
		\end{tabular}
		}
\vspace{1mm}
\end{table*}

\subsection{More Detailed Analysis} 

\noindent
\textbf{Detailed Improvements.} Table~\ref{tab:cityscapes_results_detail_val} compares the detailed results of each category on the validation set, where ResNet-101 is used as backbone, and FPN decoder with PPM head serves as the baseline. SFNet improves almost all categories, especially for 'truck' with more than 19\% mIoU improvement. Adopting GD-FAM leads to more consistent improvement over FAM on each class.

\begin{figure*}[!t]
	\centering
	\includegraphics[width=1.0\linewidth]{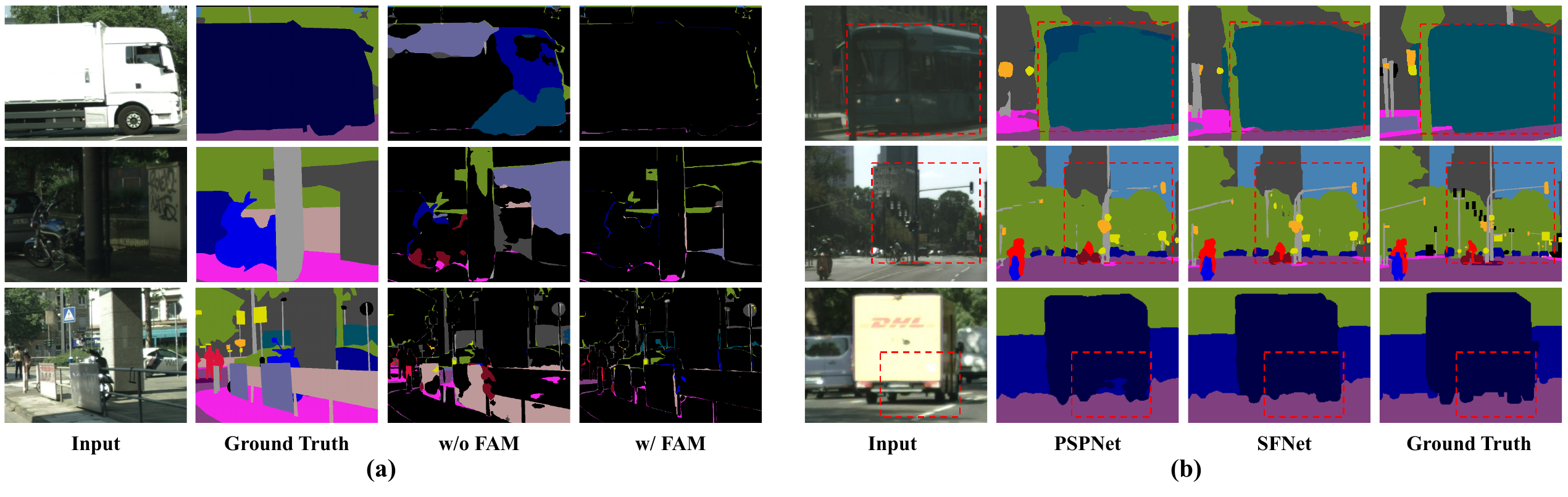}
	\caption{(a), Qualitative comparison in terms of errors in predictions, where correctly predicted pixels are shown as black background while wrongly predicted pixels are colored with their ground truth label color codes. (b), Scene parsing results comparison against PSPNet~\cite{pspnet}, where the improved regions are marked with red dashed boxes. Our method performs better on both small scale and large scale objects.}
	\label{fig:error_map}
\end{figure*}

\begin{figure*}[!t]
	\centering
	\includegraphics[width=1.0\linewidth]{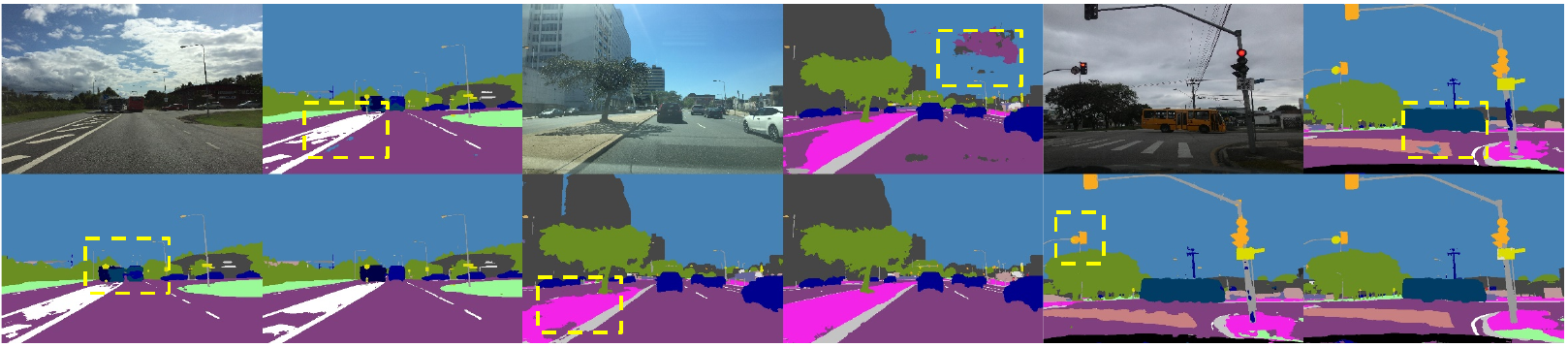}
	\caption{Qualitative comparison on Mapillary dataset. Top-left: \textit{Origin Images}. Top-Left: \textit{Results of BiSegNet~\cite{bisenet}}. Down-Left: \textit{Results of ICNet~\cite{ICnet}}. Down-Right: \textit{Results of our SFNet-Lite}. Improvement regions are in yellow boxes. Best view it in color. }
	\label{fig:mapillary_vis}
\end{figure*}

\begin{figure*}[!t]
	\centering
	\includegraphics[width=1.0\linewidth]{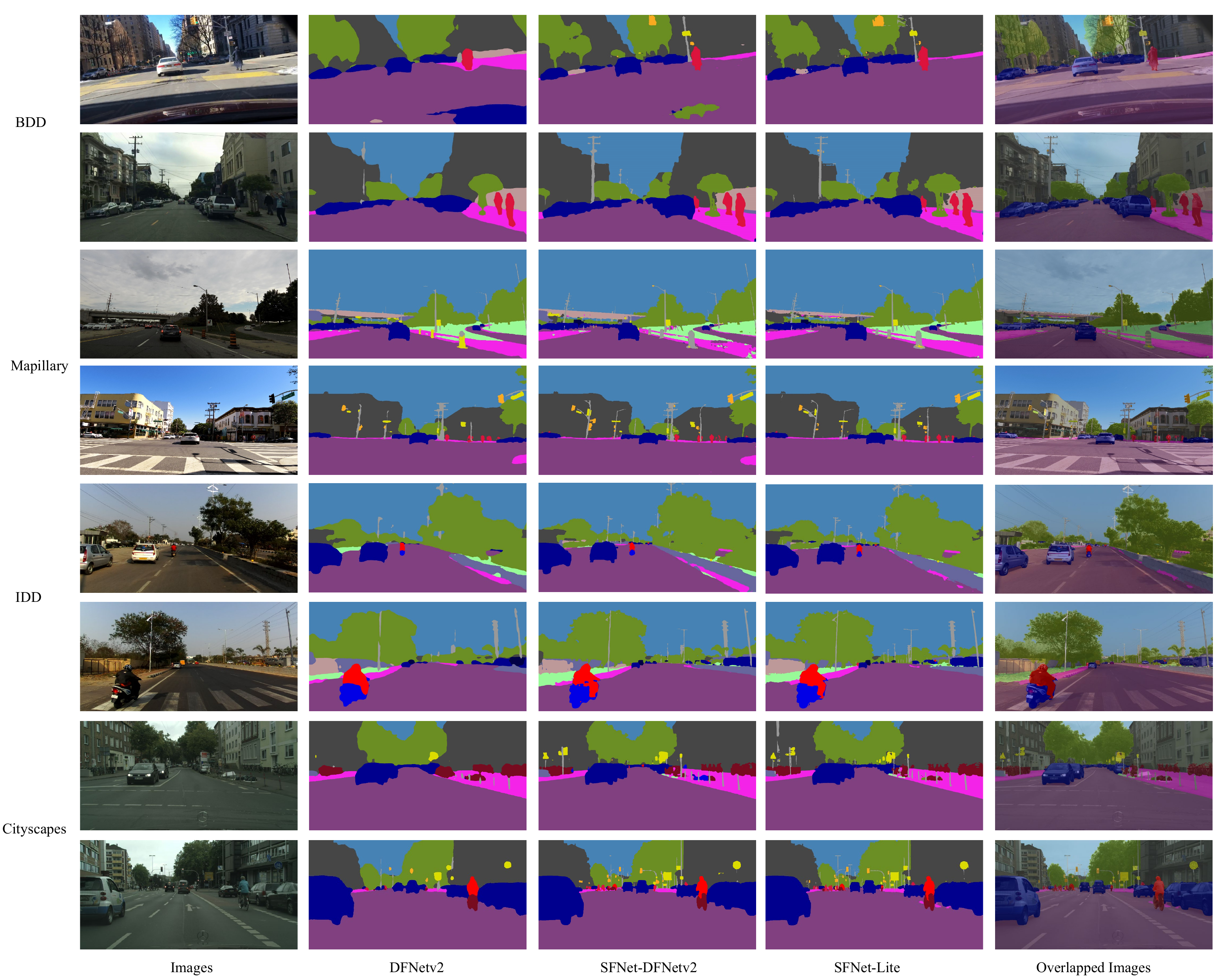}
	\caption{Visualization results on UDS validation dataset including BDD, Maillary, IDD and Cityscapes. Our methods achieve the better visual results in cases of clear object boundary, inner object consistency and better structural outputs. We adopt singele scale inference and all the models are trained under the same setting. Best view it on screen and zoom in. }
	\label{fig:res_vis_uds}
\end{figure*}

\noindent
\textbf{Visualization of Semantic Flow.} 
Figure~\ref{fig:vis_flowfield} visualizes semantic flow from FAM in different stages. Similar to optical flow, semantic flow is visualized by color coding and is bilinearly interpolated to image size for a quick overview. Besides, vector fields are also visualized for detailed inspection. From the visualization, we observe that semantic flow tends to diffuse out from some positions inside objects. These positions are generally near the object centers and have better receptive fields to activate top-level features with pure and strong semantics. Top-level features at these positions are then propagated to appropriate high-resolution positions following the guidance of semantic flow. In addition, semantic flows also have coarse-to-fine trends from the top level to the bottom level. This phenomenon is consistent with the fact that semantic flows gradually describe offsets between gradually smaller patterns.


\noindent
\textbf{Visual Improvements on Cityscapes dataset.} Figure~\ref{fig:error_map}(a) visualizes the prediction errors by both methods, where FAM considerably resolves ambiguities inside large objects (e.g., truck) and produces more precise boundaries for small and thin objects (e.g., poles, edges of wall). Figure\ref{fig:error_map} (b) shows our model can better handle the small objects with shaper boundaries than dilated PSPNet due to the alignment on lower layers.

\noindent
\textbf{Visualization Comparison on Mapillary dataset.} In Figure~\ref{fig:mapillary_vis}, we show the visual comparison results on the Mapillary dataset. As shown in that figure, compared with previous ICNet and BiSegNet, our SFNet-Lite using ResNet-18 as backbone has better segmentation results in cases of more accurate segmentation classification and structural output.

\noindent
\textbf{Visual Comparison on proposed USD dataset.} In figure~\ref{fig:res_vis_uds}, we present several samples from different datasets. Compared with the original DFNet baseline, our method can achieve better segmentation results in terms of clear object boundaries and inner object consistency. We also show the SFNet-Lite with ResNet-18 backbone in the fourth row and overlapped images in the last row. The figure shows that our methods (SFNet with DFV2 backbone and SFNet-Lite with ResNet-18 backbone) achieve good segmentation quality for different domains.

\begin{table}[!t]\setlength{\tabcolsep}{6pt}
	\centering
			\caption{Speed Comparison on TensorRT Deployment testing with Different Devices. The FPS is measured with $1024 \times 2048$ input.}
		\label{table:deployment}
	\begin{threeparttable}
		\scalebox{0.90}{
			\begin{tabular}{l c c c}
				\toprule[0.2em]
				Network & 1080-TI & TITAN-RTX & RTX-3090 \\  
				\toprule[0.2em]
				SFNet(resnet-18)& 26.8 & 34.2  & 50.5  \\
				SFNet(stdcv2) & 56.2 &  78.0 &  202.1  \\
				\bottomrule[0.1em]
			\end{tabular}
		}
	\end{threeparttable}
\end{table}

\noindent
\textbf{Speed Effect on Different Devices.} In Table~\ref{table:deployment}, we explore the effect of deployment devices. In particular, compared with the original SFNet~\cite{SFnet}, which uses 1080-TI as a device, using a more advanced device leads to a much higher speed. For example, RTX-3090 results almost twice faster as 1080-TI using ResNet-18 and four times faster using STDCNet. Moreover, we also find that SFNet with STDCNet~\cite{STDCNet} backbone is more friendly to TensorRT deployment.


\begin{table}[!t]\setlength{\tabcolsep}{6pt}
	\centering
			\caption{Pretraining Effect of UDS dataset. The mIoU is evaluated on Camvid dataset~\cite{CamVid}.}
		\label{table:effect_pretraining}
	\begin{threeparttable}
		\scalebox{0.90}{
			\begin{tabular}{l c c c}
				\toprule[0.2em]
				Network & ImageNet & UDS & mIoU \\  
				\toprule[0.2em]
				SFNet(resnet-18)&  \checkmark & - & 73.8   \\
				SFNet(stdcv2) &  \checkmark &  - & 72.9 \\
				\midrule
				SFNet(resnet-18)&  -  & \checkmark  & 76.5 \\
			    SFNet(stdcv2) & - & \checkmark &  75.6 \\
				\bottomrule[0.1em]
			\end{tabular}
		}
	\end{threeparttable}
\end{table}

\noindent
\textbf{UDS Used for Pre-training.} We further show the effectiveness of our UDS dataset in table~\ref{table:effect_pretraining}. Compared with ImageNet~\cite{imagenet}, adopting the pre-training with the UDS dataset can significantly boost SFNet results on the Camvid dataset~\cite{CamVid}, which leads to a significant margin (3-4\% mIoU). This implies that the UDS dataset can be an excellent pre-train source to boost the model performance.

\subsection{Extension on Efficient Panoptic Segmentation}

\noindent
\textbf{Experiment Setting.} {In this section, we show the generalization ability of our Semantic Flow on more challenging task Panoptic Segmentation. We choose K-Net~\cite{zhang2021knet} as the prediction head, while our SFNet is the backbone and neck for the feature extractor. All the network is first trained on the COCO dataset and then on the Cityscapes dataset. For COCO~\cite{coco_dataset} dataset pretraining, all the models are trained following detectron2 settings~\cite{detectron2}. We adopt the multiscale training by resizing the input images such that the shortest side is at least 480 and 800 pixels, while the longest is at most 1333 pixels. We also apply random crop augmentations during training, where the training images are cropped with a probability of 0.5 to a random rectangular patch and then resized again to 800-1333 pixels. All the models are trained for 36 epochs. For Cityscape fine-tuning,  we resize the images with a scale ranging from 0.5 to 2.0 and randomly crop the whole image during training with batch size 16. All the results are obtained via single-scale inference. We also report results using the ResNet50 backbone for reference. We report the FPS on V100 devices by averaging 100 input images. For FPS measurement, we also include the panoptic post-processing times.}

\begin{table}
\centering
\caption{{{Experiment results on the Cityscapes Panoptic validation set.} $^*$ indicates using DCN~\cite{deformable}. All the methods use single-scale inference. We prove the generalization ability of semantic flow. The FPS is measured on one V100 card with $1024\times2048$ input.}}
\label{tab:results_city_panoptic}
\begin{adjustbox}{width=0.50\textwidth}
\begin{tabular}{c c c c c c}
\toprule[0.15em]
\textbf{Method}  &  Backbone  & $PQ$ & $PQ_{th}$ & $PQ_{st}$ & \#FPS  \\ 
\midrule[0.15em]
UPSNet~\cite{xiong2019upsnet} & ResNet50 & 59.3 & 54.6 & 62.7 & 7.3 \\
SOGNet~\cite{yang2019sognet} & ResNet50 & 60.0 & {56.7} & 62.5 & 6.7 \\
Seamless~\cite{porzi2019seamless} & ResNet50 & 60.2 & 55.6 & 63.6 & -\\
Unifying~\cite{li2020unifying} & ResNet50 & 61.4 & 54.7 & 66.3 & -\\
Panoptic-DeepLab~\cite{cheng2020panoptic} & ResNet50 & 59.7 & - & - & 8.2 \\
Panoptic FCN$^*$~\cite{li2020panopticFCN} & ResNet50 & 61.4 & 54.8 & 66.6 & - \\
K-Net~\cite{zhang2021knet} & ResNet50 & 61.2 & 52.4 & 66.8 & 10.2 \\
\hline
STDCv1 + K-Net Head  & STDCv1 & 58.0 &  50.3 & 62.4 & 23.3  \\
SF-STDCv1 + K-Net Head &  STDCv1 & 59.2  & 52.9 & 63.3 & 20.3 \\
\hline
STDCv2 + K-Net Head   & STDCv2 &  59.8 & 53.8 & 63.8 & 19.3 \\
SF-STDCv2 + K-Net Head & STDCv2 & 60.3  &  54.4 &  64.8 & 18.6 \\
\hline
SF-ResNet50 + K-Net Head & ResNet50 & 61.7 & 52.6 & 67.2 & 9.0 \\
\bottomrule
\end{tabular}
\end{adjustbox}

\end{table}

\noindent
\textbf{Results on Various Baseline on Cityscapes Panoptic Segmentation.} {As shown in Table~\ref{tab:results_city_panoptic}, our SFNet backbone improves the baseline models in terms of the Panoptic Quality metric by around 0.5-1.0\%. The results show the generalization ability of the semantic flow because our aligned feature representation preserves more fine-grained information. Moreover, we compare our methods using a stronger ResNet50 backbone. Compared with K-Net~\cite{zhang2021knet}, our methods still achieve 0.5\% PQ improvements with 1.2 FPS drop. Our method with STDCv2 backbone achieves a strong speed and accuracy trade-off (60.3 PQ with 18.6 FPS).}

\subsection{More Analysis on SFNet and SFNet-Lite}

\noindent
\textbf{Experiment Setting.} {In this section, we perform more extensive experiments using SFNets. (1), We first conduct more experiments with DCN~\cite{deformable} using Cityscapes and UDS by adding one DCN layer and one GD-FAM. (2), Then, we perform domain generalization experiments using RobustNet with different SFNet baselines, where we train the model on the Cityscapes dataset and test the model on BDD and IDD datasets. (3), Next, we present the results on ADE20k datasets using different baselines, including Semantic FPN~\cite{PanopticFPN} and SegFormer~\cite{xie2021segformer}. For the experiments on the ADE20k dataset, we follow the default settings from OCRNet~\cite{OCRnet}, where the crop size is set to 512 with 160k iterations training. The GFlops are calculated with $512 \times 512$ inputs.}

\begin{table}[!t]\setlength{\tabcolsep}{6pt}
	\centering
	\caption{{More detailed comparison between GF-FAM and DCN. We adopt ResNet18 as backbone.}}
	\label{table:more_comparison_dcn}
	\begin{threeparttable}
		\scalebox{0.90}{
			\begin{tabular}{l c c c}
				\toprule[0.2em]
				Method & Cityscapes & UDS & \#FPS \\  
				\toprule[0.2em]
				FCN +FPN + PPM (baseline) &  76.6 & 72.3  & 40.3 \\
				w one DCN  &  76.9 & 73.5 & 22.8  \\
				w one FAM & 77.5 &  74.8 & 23.3 \\
				w one GD-FAM  & 78.5  & 75.5  & 24.6  \\
				\bottomrule[0.1em]
			\end{tabular}
		}
	\end{threeparttable}
\end{table}

\noindent
\textbf{More Detailed Comparison with DCN.} {We carry out a more detailed comparison between DCN and our proposed GD-FAM. In particular, we replace GD-FAM or FAM with a simple concatenation followed by a deformable convolution, where GD-FAM and FAM are inserted in the last stage to align the last two features for comparison. The DCN directly replaces FAM or GD-FAM. As shown in Tab~\ref{table:more_comparison_dcn}, our method achieves better results (1.0-2.0\% mIoU gains) on both Cityscape dataset and UDS dataset, which share the same conclusion with the findings in Tab.~\ref{tab:city_ablation_FAM} (d). 
}
\begin{table}[!t]\setlength{\tabcolsep}{6pt}
	\centering
\caption{{Domain generalization experiments using SFNet and SFNet-Lite using RobustNet~\cite{choi2021robustnet}. The baseline methods are SFNet series with RobustNet with no FAMs or GD-FAMs. Our methods also show better results on BDD and IDD when trained with Cityscapes dataset.}}
	\label{table:dg_exp}
	\begin{threeparttable}
		\scalebox{0.90}{
			\begin{tabular}{l c c c c}
				\toprule[0.2em]
				Method &  Backbone & BDD & IDD & \#FPS \\  
				\toprule[0.2em]
			     baseline & ResNet18 & 43.2  & 46.2 & 25.8 \\ 
			     SFNet & ResNet18 &  45.2 & 48.1 & 24.0 \\
			    SFNet-Lite & ResNet18 & 46.0 & 48.5 & 24.8 \\
			    \hline
			     baseline & STDC-2 & 39.2 & 41.4 & 27.5 \\ 
			     SFNet-Lite & STDC-2 & 41.8 & 44.5 & 26.7 \\
				\bottomrule[0.1em]
			\end{tabular}
		}

	\end{threeparttable}
\end{table}

\noindent
\textbf{Domain Generalization Testing Using RobustNet~\cite{choi2021robustnet}.} {We further prove the domain generalization ability of SFNet and SFNet-Lite. Our methods are based on previous work RobustNet~\cite{choi2021robustnet} and Semantic-FPN~\cite{PanopticFPN}. In particular, we follow the original open-source RobustNet code~\footnote{\url{https://github.com/shachoi/RobustNet}} and settings by the whitening operation in different backbones to build the baseline. As shown in Tab~.\ref{table:dg_exp}, our methods achieve consistent 2-3\% mIoU improvements over the RobustNet baselines on both IDD and BDD datasets.
}

\begin{table}[!t]\setlength{\tabcolsep}{6pt}
	\centering
	\caption{{Effectiveness on ADE20k dataset on Semantic-FPN with different backbones.}}
    \label{table:ade_20k_cnn}
	\begin{threeparttable}
		\scalebox{0.90}{
			\begin{tabular}{l c c c c}
				\toprule[0.2em]
				Method &  Backbone & mIoU & \#GFLOP \\  
				\toprule[0.2em]
			    Semantic-FPN (baseline) & ResNet50 & 37.6 & 72.8 \\ 
			    SFNet & ResNet50  & 39.0 & 75.2 \\
			     SFNet-Lite & ResNet50  & 38.8 & 74.2 \\
			     \hline
			    Semantic-FPN (baseline) & DF2 & 34.5 & 35.2 \\ 
			     SFNet & DF2 & 36.7 & 36.2 \\
			     SFNet-Lite & DF2 & 36.8 & 35.9 \\
				\bottomrule[0.1em]
			\end{tabular}
		}
	\end{threeparttable}
\end{table}

\begin{table}[!t]\setlength{\tabcolsep}{6pt}
	\centering
	\caption{{Effectiveness on ADE20k dataset on Transformer-based methods. B0 and B1 : backbones in SegFormer~\cite{xie2021segformer}.}}
	\label{table:ade_20k_transformer_based}
	\begin{threeparttable}
		\scalebox{0.90}{
			\begin{tabular}{l c c c}
				\toprule[0.2em]
				Method &  Backbone & mIoU & \#GFLops \\  
				\toprule[0.2em]
			    SegFormer (baseline) & B0 & 37.4 & 8.4  \\ 
			     SFNet-Lite & B0 & 38.2 & 9.4 \\
			     \hline
			     SegFormer (baseline) & B1 & 40.9 & 16.0 \\ 
			     SFNet-Lite & B1 & 42.2 & 17.6 \\
				\bottomrule[0.1em]
			\end{tabular}
		}
	\end{threeparttable}
\end{table}

\noindent
\textbf{Experiment Results on ADE20k Dataset.} { In Tab.~\ref{table:ade_20k_cnn}, we verify the effectiveness of FAM and GD-FAM on the more challenging dataset ADE20k. For a fair comparison, we re-implement the baseline in the same codebase and report our reproduced results for Semantic-FPN. As shown in that table, we find about 1.2\%-2.2\% improvements over different baselines. In particular, we find the improvements on the real-time model are stronger, which means the semantic gaps in small models are heavier. This finding is similar in the road driving scene datasets (see Tab.~\ref{table:bdd_sota_speed_acc}, Tab.~\ref{table:uds_sota_speed_acc}).
}

\noindent
\textbf{Experiment Results on ADE20k Using Transformer-based Model.} { In Tab.~\ref{table:ade_20k_transformer_based}, we also report the results using transformer-based model SegFormer~\cite{xie2021segformer}. We also find about 0.8-1.3\% mIoU improvements over different backbones. These results indicate that our proposed approach can also be used in transformer-based segmenter.}

\section{Conclusion}
In this paper, we propose to use the learned \textbf{Semantic Flow} to align multi-level feature maps generated by aligned feature pyramids for semantic segmentation. We propose a flow-aligned module to fuse high-level feature maps and low-level feature maps. Moreover, to speed up the inference procedure, we propose a novel Gated Dual flow alignment module to align both high and low-resolution feature maps directly. By discarding atrous convolutions to reduce computation overhead and employing the flow alignment module to enrich the semantic representation of low-level features, our network achieves the best trade-off between semantic segmentation accuracy and running time efficiency. Experiments on multiple challenging datasets illustrate the efficacy of our method. Moreover, we merge four challenging driving datasets into one Unified Driving Segmentation dataset (UDS), which contains various domains. We benchmark several works on the merged dataset. Experiment results show that our SFNet series can achieve the best speed and accuracy trade-off. In particular, our SFNet improves the original DFNet on the UDS dataset by a large margin (9.0\% mIoU). These results indicate that our SFNet can be a faster and accurate baseline for Semantic Segmentation.

\noindent
\textbf{Acknowledgement} 
We gratefully acknowledge the support of SenseTime Research for providing the computing resources to carry out this research. Without it, we could not perform benchmarking. This work was supported by the National Key Research and Development Program of China (No.2020YFB2103402). We also thank Qi Han (SenseTime) and Houlong Zhao (XiaoMi Car) for discussing the TensorRT deployment.

\noindent
\textbf{Data Availability Statement}
All the datasets used in this paper are available online. Cityscapes~\footnote{\url{https://www.cityscapes-dataset.com/benchmarks/}}, BDD~\footnote{\url{https://bdd-data.berkeley.edu/}}, IDD~\footnote{\url{https://idd.insaan.iiit.ac.in/}}, Mapillary~\footnote{\url{https://www.mapillary.com/dataset/vistas}}, and Camvid~\footnote{\url{http://mi.eng.cam.ac.uk/research/projects/VideoRec/CamVid/}} can be downloaded from their official website accordingly. The UDS dataset is merged from these datasets. 

{\small
\bibliographystyle{unsrt}
\bibliography{reference}}

\end{document}